\documentclass{sig-alt-full}

\usepackage[pdftex]{hyperref}
\hypersetup{
pdfauthor={Arun Kejariwal}, pdfcreator={pdflatex},
pdfproducer={pdflatex}, pdfsubject={SIGMETRICS 2007 Paper},
pdfkeywords={Anomalies, Cloud Computing, BigData, Performance},
pdfborder={0 0 0},
pdfpagemode=UseOutlines, 
colorlinks=true,
anchorcolor = JungleGreen,
filecolor = cyan,
menucolor=blue,
pagecolor = magneta,
citecolor = Magenta, 
linkcolor=red, 
urlcolor=blue,
bookmarks=true,
backref=true,
bookmarksopen,
bookmarksnumbered,
hyperindex=true
}

\usepackage[dvipsnames,usenames]{xcolor}
\definecolor{webgreen}{rgb}{0, 0.5, 0} 
\definecolor{webblue}{rgb}{0, 0, 0.5} 
\definecolor{webred}{rgb}{0.5, 0, 0} 

\usepackage{pifont}
\renewenvironment{dinglist}[2][Blue]
{\begin{list}{\textcolor{#1}{\ding{#2}}}{}}{\end{list}}
\DeclareFontFamily{OT1}{pzc}{}%
\DeclareFontShape{OT1}{pzc}{m}{it}{<-> s * [1.200] pzcmi7t}{}%
\DeclareMathAlphabet{\mathpzc}{OT1}{pzc}{m}{it}

\usepackage{eucal}
\usepackage{mathrsfs}

\usepackage{subcaption}
\usepackage[pdftex]{floatflt}
\usepackage{graphicx}
\usepackage{multirow}
\usepackage{latexsym}

\usepackage{amssymb}
\usepackage{amsmath}
\usepackage{shadow}
\usepackage{amsbsy}
\usepackage{amsfonts}
\usepackage{enumerate}
\usepackage{moreverb}
\usepackage{array}

\usepackage{fancyhdr}
\usepackage{colortbl}
\usepackage{xcolor}
\newcommand\BlackCell[1]{%
  \multicolumn{1}{|c|}{\cellcolor{lightgray}\textcolor{black}{#1}}
}

\newlength{\additionalvspace}%
\newlength{\colwidth}%
\newlength{\colwidthA}%
\newlength{\colwidthB}%
\newlength{\colwidthBb}%
\newlength{\colwidthC}%
\usepackage{algorithm}
\usepackage{algorithmic}
\renewcommand{\algorithmiccomment}[1]{/* #1 */}

\def\figref#1{Figure~\ref{#1}}
\def\secref#1{Section~\ref{#1}}
\def\ssecref#1{subsection~\ref{#1}}
\def\sssecref#1{subsubsection~\ref{#1}}

\def\eqnref#1{Equation~\ref{#1}}
\def\tabref#1{Table~\ref{#1}}
\def\algoref#1{Algorithm~\ref{#1}}

\newtheorem{theorem}{Theorem}

\newtheorem{definition}[theorem]{Definition}

\usepackage{sty/dblfloatfix}
\usepackage{wasysym}

\graphicspath{{figs/}}

\title{Automatic Anomaly Detection in the Cloud \\ Via Statistical Learning}

\author{
\begin{tabular}{c c c}
 Jordan Hochenbaum & Owen S. Vallis & Arun Kejariwal 
\end{tabular} \\[1mm]
\begin{tabular}{c}
Twitter Inc. \\
\end{tabular}
}

\newcommand*{\refname}{Bibliography}
\begin{document}

\maketitle

\thispagestyle{empty}

\begin{abstract}
{\em

Performance and high availability have become increasingly important drivers, 
amongst other drivers, for user retention in the context of web services such 
as social networks, and web search. Exogenic and/or endogenic factors 
often give rise to anomalies, making it very challenging to maintain high 
availability, while also delivering high performance. Given that service-oriented 
architectures (SOA) typically have a large number of services, with each 
service having a large set of metrics, automatic detection of anomalies 
is non-trivial. 
Although there exists a large body of prior research in anomaly detection, 
existing techniques are not applicable in the context of social network data,
owing to the inherent seasonal and trend components in the time series data.

To this end, we developed two novel statistical techniques for automatically
detecting anomalies in cloud infrastructure data. Specifically, the techniques
employ statistical learning to detect anomalies in both application, and system
metrics. Seasonal decomposition is employed to filter the trend
and seasonal components of the time series, followed by the use of robust statistical metrics -- 
median and median absolute deviation (MAD) -- to accurately detect anomalies,
even in the presence of seasonal spikes.  
We demonstrate the efficacy of the proposed techniques from three different
perspectives, viz., capacity planning, user behavior, and supervised learning. 
In particular, we used \underline{production} data for evaluation, and we report
Precision, Recall, and F-measure in each case. 

}
 
\end{abstract}

\section{Introduction} \label{sec:intro}

Big Data is characterized by the increasing volume (on the order of zetabytes), 
and the velocity of data generation \cite{NSFBigData,Manyika11}. It is projected
that the market size of Big Data shall climb up from the current market size of 
\$5.1 billion \cite{BigDataMarket} to \$53.7 billion by 2017. In a recent 
report \cite{EMCReport}, EMC Corporation stated: {\em ``A major factor behind
the expansion of the digital universe is the growth of machine generated data,
increasing from 11\% of the digital universe in 2005 to over 40\% in 2020."}
In the context of social networks, analysis of User Big Data is key to building
an engaging social network; in a similar fashion, analysis of Machine Big Data 
is key to building an efficient and performant underlying cloud computing 
platform.

Anomalies in Big Data can potentially result in losses to the business -- in both 
revenue \cite{AWSOutage}, as well as in long term reputation \cite{Repute}. To this
end, several enterprise-wide monitoring initiatives \cite{Agarwala06,Ren10} have
been undertaken. Likewise, there has been an increasing emphasis on developing 
techniques for detection, and root cause analysis, of performance issues in the 
cloud \cite{Chen02,Babenko09,Magalhaes11,Kang12,Attariyan12,Wang10,Wang11}. 

A lot of research has been done in the context of anomaly detection in various 
domains such as, but not limited to, statistics, signal processing, finance,
econometrics, manufacturing, and networking \cite{hawkins_identification_1980,
barnett_outliers_1994,hodge_survey_2004,aggarwal_outlier_2013}. In a recent 
survey paper Chandola et al.\ highlighted that anomalies are contextual in 
nature \cite{Chandola09} and remarked the following: 

\begin{quote} 
{\em 
A data instance might be a contextual anomaly in a given context, but an 
identical data instance (in terms of behavioral attributes) could be
considered normal in a different context. This property is key in identifying
contextual and behavioral attributes for a contextual anomaly detection 
technique.
}
\end{quote} 

\noindent
Detection of anomalies in the presence of seasonality, and an underlying trend -- 
which are both characteristic of the time series data of social networks -- is 
non-trivial. \figref{fig:intro} illustrates the presence of both positive and
negative anomalies -- corresponding to the circled data points -- in time 
series data obtained from production. From the figure we note that the time 
series has a very conspicuous seasonality, and that there are multiple modes 
within a seasonal period. Existing techniques for anomaly detection 
(overviewed in-depth in \secref{rel}) are not amenable for time series data 
with the aforementioned characteristics. To this end, we developed novel techniques 
for automated anomaly detection in the cloud via statistical learning. In 
particular, the main contributions of the paper are as follows:

\begin{figure*}[!t]
\centering 
\includegraphics[width=\linewidth,height=1.8in]{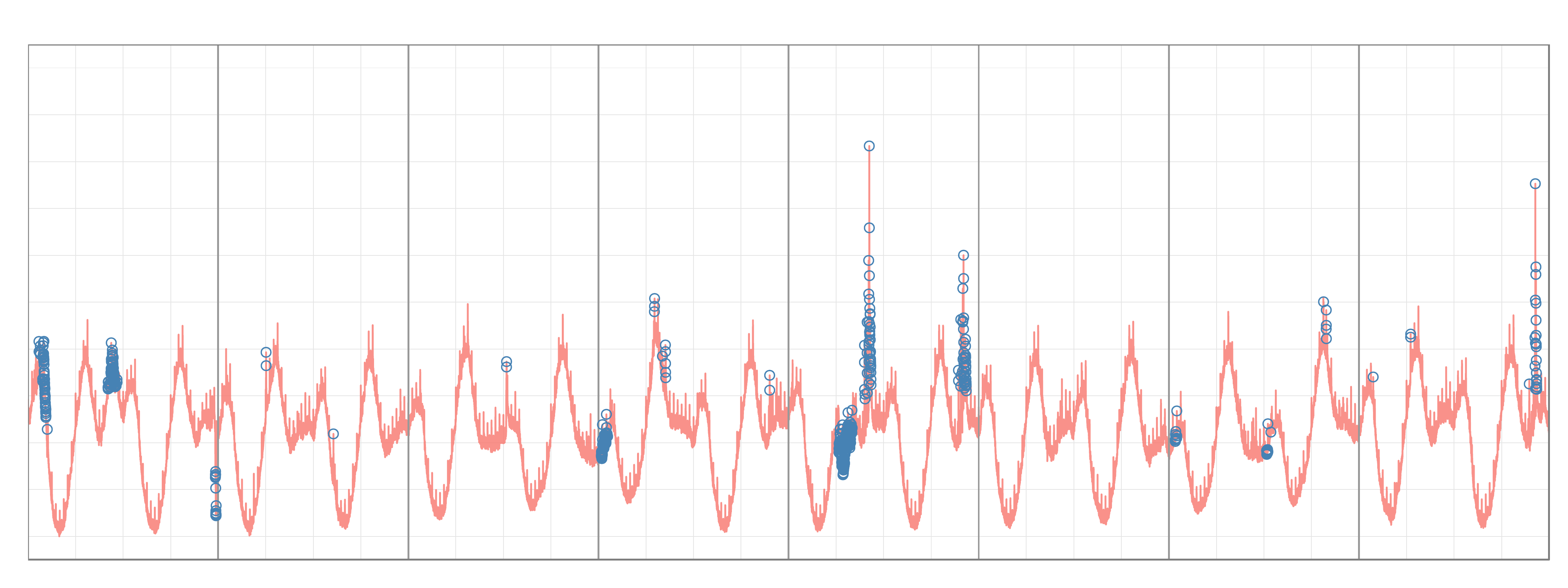}
\caption{An illustrative time series, obtained from \underline{production}, with positive and negative anomalies}
\label{fig:intro}
\end{figure*}

\begin{dinglist}{114}
\item First, we propose novel statistical learning based techniques to detect
      anomalies in the cloud. The proposed techniques can be used to automatically
      detect anomalies in time series data of both application metrics such as 
      Tweets Per Sec (TPS) and system metrics such as CPU utilization etc. 
      Specifically, we propose the following: 
      \begin{dinglist}{122}
      \item {\bf Seasonal ESD (S-ESD)}: This techniques employs time series 
            decomposition to determine the seasonal component of a given time
            series. {\em S-ESD} then applies ESD \cite{rosner_detection_1975,
            rosner_percentage_1983} on the resulting time series to detect 
            the anomalies.  
      \item {\bf Seasonal Hybrid ESD (S-H-ESD)}: In the case of some time series 
            (obtained from \underline{production}) we observed a relatively 
            high percentage of anomalies. To address such cases, coupled with  
            the fact that mean and standard deviation (used by ESD) are highly
            sensitive to a large number anomalies \cite{leys_detecting_2013,snedecor_statistical_1989}, 
            we extended {\em S-ESD} to use the robust statistics {\em median} 
            \cite{huber_robust_1981} and median absolute deviation (MAD) to 
            detect anomalies \cite{hampel_influence_1974}. Computationally, 
            {\em S-H-ESD} is more expensive than {\em S-ESD}, but is more robust
            to a higher percentage of anomalies.  
      \end{dinglist}
\item Second, we present a detailed evaluation of the proposed techniques using
      production data. In particular, we present the evaluation from three 
      different perspectives: 
      \begin{dinglist}{122}
      \item {\em Capacity planning}: Occurrence of anomalies can potentially 
            result in violation of service level agreement (SLA), and/or impact 
            end-user experience. To mitigate the impact of anomalies, timely 
            detection of anomalies is paramount and may require provisioning 
            additional capacity. Given a threshold for a system metric 
            (corresponding to the SLA), we evaluate the efficacy of the  
            proposed techniques to detect anomalies greater than the specified 
            threshold.  
      \item {\em User behavior}: Change in user behavior -- which may result 
            due to, but not limited to, events such as the Superbowl -- at 
            times manifests itself as anomalies in time series data. Thus,
            anomaly detection can guide the study of change in user behavior.
            To this end, we evaluated the efficacy of the proposed techniques
            with respect to change in user behavior.  
      \item {\em Supervised Learning}: Given the velocity, volume, and real-time
            nature of cloud infrastructure data, it is not practical to obtain 
            time series data with ``true" anomalies labeled. To address this  
            limitation, we injected anomalies in a randomized fashion in smoothed,
            using B-spline, production data. The randomization was done along three 
            dimensions -- time of injection, magnitude, and width of the anomaly. 
            We evaluated the efficacy of the proposed techniques with respect to 
            detection of the injected anomalies.  
      \end{dinglist}
\end{dinglist}

\noindent

The rest of the paper is organized as follows: \secref{sec:background} lays out
the notation used in the rest of the paper, and briefly overviews statistical 
background for completeness and better understanding of the rest of the paper. 

\secref{sec:techniques} details the techniques proposed in this paper, viz., 
{\bf S-ESD} and {\bf S-H-ESD}. \secref{sec:eval} presents a detailed evaluation
of the aforementioned techniques using \underline{production} data. 

Previous work is discussed in \secref{rel}. Finally, in \secref{sec:conclusions}
we conclude with directions for future work.

\section{Background} \label{sec:background}

In this section we describe the notation used in the rest of the paper.
Additionally, we present a brief statistical background for completeness,
and for better understanding of the rest of the paper. 

A time series refers to a set of observations collected sequentially 
in time. Let $x_t$ denote the observation at time $t$, where $t = 0, 1, 2,
\ldots$, and let $X$ denote the set of all observations constituting the time series.

\subsection{Grubbs Test and ESD} \label{sec:outliertests} 

In this subsection, we briefly overview the most widely used existing techniques for
anomaly detection. In essence, these techniques employ statistical hypothesis 
testing, for a given significance level \cite{Fisher25}, to determine whether 
a datum is anomalous. In other words, the test is used to evaluate the rejection 
of the {\em null hypothesis} ($H_0$), for a pre-specified level of significance, 
in favor of the {\em alternative hypothesis} ($H_1$).

\subsubsection{Grubbs Test} \label{sec:grubbs} 

Grubbs test \cite{Grubbs50,grubbs_procedures_1969} was developed for detecting 
the largest anomaly within a univariate sample set. The test assumes that the 
the underlying data distribution is normal. 
Grubbs' test is defined for the hypothesis:

\begin{align}
H_0: & \text{There are no outliers in the data set} \\ 
H_1: & \text{There is at least one outlier in the data set} 
\end{align}

\noindent
The Grubbs' test statistic is defined as follows:

\begin{equation} C = \frac{\max_{t} \mid x_{t} - \overline{x} \mid}{s} \end{equation}

\noindent
where, $\overline{x}$ and $s$ denote the mean and variance of the time series X. 
For the two-sided test, the hypothesis of no outliers is rejected at
significance level $\alpha$ if

\begin{equation} 
C > \frac{(N - 1)}{\sqrt{N}}\sqrt{\frac{(t_{\alpha/(2N),N-2})^{2}}{N - 2 + (t_{\alpha/(2N),N-2})^{2}}} 
\label{eq:grubb}
\end{equation}
where $t_{\alpha/(2N),N-2}$ denotes the upper critical values of the t-distribution with $N-2$ degrees of freedom and a significance level of $\alpha/(2N)$. For one-sided tests, $\alpha/(2N)$ becomes $\alpha/N$ \cite{garcia_tests_2012}.
\noindent
The largest data point in the time series that is greater than the test statistic 
is labeled as an anomaly. 

In practice, we observe that there is more than one anomaly in the time series
data obtained from production. Conceivably, one can iteratively apply Grubbs'
test to detect multiple anomalies. Removal of the largest anomaly at each 
iteration reduces the value of $N$; however, Grubbs' test does not update the
value obtained from the t-distribution tables. Consequently, Grubbs' test is
not suited for detecting multiple outliers in a given time series data. 

Several other approaches, such as the Tietjen-Moore test\footnote{Although the
Tietjen-Moore test can be used to detect multiple anomalies, it requires the 
number of anomalies to detect to be pre-specified. This is not practical in the
current context.} \cite{tietjen_grubbs-type_1972}, and the extreme Studentized 
deviate (ESD) test \cite{rosner_detection_1975,rosner_percentage_1983} have 
been proposed to address the aforementioned issue.  Next, we briefly overview 
ESD.

\subsubsection{Extreme Studentized Deviate (ESD)} \label{sec:esd}

The Extreme Studentized Deviate test (ESD) \cite{rosner_detection_1975} (and its 
generalized version \cite{rosner_percentage_1983}) can also be used to detect 
multiple anomalies in the given time series. Unlike the Tietjen-Moore test, it 
only requires an upper bound on the number of anomalies ($k$) to be specified. 
In the worst case, the number of anomalies can be at most 49.9\% of the total 
number of data points in the given time series. In practice, our observation, 
based on production data, has been that the number of anomalies is typically 
less than $1\%$ in the context of application metrics and less than $5\%$ in the 
context of system metrics.  

ESD computes the following test statistic for the $k$ most extreme values 
in the data set.

\begin{equation} 
C_{k}=\frac{max_{k}\mid x_{k} - \overline{x}\mid}{s}
\label{eq:esd-test-stat}
\end{equation}

\noindent 
The test statistic is then compared with a critical value, computed using 
\eqnref{eq:esd-critical}, to determine whether a value is anomalous. If 
the value is indeed anomalous, it is removed from the data set, and the 
critical value is recalculated from the remaining data. 

\begin{equation} 
\lambda_{k}=\frac{(n - k)t_{p}, _{n - k - 1}}{\sqrt{(n - k - 1 + t_{p}^{2}, _{n - k - 1})(n - k + 1)}} 
\label{eq:esd-critical}
\end{equation}

\noindent 
ESD repeats this process $k$ times, with the number of anomalies equal to the
largest $k$ such that $C_{k} > \lambda_{k}$. In practice, $C_{k}$ may swing
above and below $\lambda_{k}$ multiple times before permanently becoming less
then $\lambda_{k}$. 

In the case of Grubbs' test, the above would cause the test to prematurely
exit; however, ESD will continue until the test has run for $k$ outliers.

\subsection{Median and Median Absolute Deviation} \label{sec:robuststats} 

The techniques discussed in the previous subsection use mean and standard deviation. It is well known that these metrics are sensitive to anomalous data \cite{huber_robust_1981, hampel_robust_1986}. As such, the use of the statistically robust median, and the median absolute deviation (MAD), has been proposed to address these issues. 

The sample mean $\bar{x}$ can be distorted by a single anomaly, with the distortion 
increasing as $x_{t} \to \pm \infty$. By contradistinction, the sample median is robust 
against such distortions, and can tolerate up to 50\% of the data being anomalous. Thus, the 
sample mean is said to have a {\em breakdown point} \cite{hampel_contributions_1968,donoho_notion_1983} 
of 0, while the sample median is said to have a {\em breakdown point} of 0.5.

For a univariate data set $X_{1}, X_{2}, ..., X_{n}$, MAD is defined as the median 
of the absolute deviations from the sample median. Formally,

\begin{equation} 
MAD = median_{i}(\left | X_{i}- median_{j}(X_{j})\right |) 
\end{equation}

\noindent 
Unlike standard deviation, MAD is robust against anomalies in the input data \cite{
hampel_influence_1974,huber_robust_1981}. Furthermore, MAD can be used to estimate
standard deviation by scaling MAD by a constant factor $b$. 

\begin{equation}
\hat{\sigma}=b\cdot  MAD
\end{equation}

\noindent 
where $b = 1.4826$ is used for normally distributed data (irrespective of non-normality
introduced by outliers) \cite{rousseeuw_alternatives_1993}. When another 
underlying distribution is assumed, Leyes et al.\ suggest $b=\frac{1}{Q(0.75)}$, 
where Q(0.75) is the 0.75 quantile of the underlying distribution 
\cite{leys_detecting_2013}.

\subsection{Precision, Recall, and F-Measure} \label{sec:fmeasure}

As mentioned earlier in \secref{sec:intro}, the efficacy of the proposed techniques 
were evaluated from three different perspectives. In each case we report the following 
metrics -- {\em Precision, Recall}, and {\em F-measure} (these metrics are commonly 
used to report the efficacy of an algorithm in data mining, information retrieval, 
et cetera). 

In the context of anomaly detection, {\em Precision} is defined as follows: 

\begin{equation} 
\text{Precision} = \frac{\vert \lbrace S \rbrace \cap \lbrace G \rbrace \vert}{\vert \lbrace S \rbrace \vert} = \frac{tp}{tp+fp} 
\label{eq:prec} 
\end{equation} 

\noindent 
where $S$ is the set of detected anomalies, and $G$ is the set of ground-truth
anomalies. In other words, {\em Precision} is the ratio of true positives (tp)
over the sum of true positives (tp) and false positives (fp)
that have actually been detected. 

In the context of anomaly detection, {\em Recall} is defined as follows: 
 
\begin{equation} 
\text{Recall} = \frac{\vert \lbrace S \rbrace \cap \lbrace G \rbrace \vert}{\vert \lbrace G \rbrace \vert} = \frac{tp}{tp+fn} 
\end{equation} 

\noindent
where {\em fn} denotes false negatives. Lastly, {\em F-measure} is defined as follows: 

\begin{equation} 
F = 2 \times \frac{\text{precision} \times \text{recall}}{\text{precision} + \text{recall}} 
\label{eq:fm}
\end{equation}

\noindent 
Since Precision and Recall are weighted equally in \eqnref{eq:fm}, the measure
is also referred as the $F_{1}$-measure, or the balanced F-score. Given that 
there is a trade-off between {\em Precision} and {\em Recall}, the {\em F-measure}
is often generalized as follows:

\begin{equation*} 
F_{\beta} = (1+\beta^{2}) \times \frac{\text{precision} \times \text{recall}}{\beta^{2} \times \text{precision} + \text{recall}}\ \ \text{where}\ \beta \geq 0
\end{equation*}

\noindent
If $\beta >1$, F is said to become more recall-oriented and if $\beta < 1$, F is 
said to become more precision-oriented \cite{sasaki_truth_2007}.

\begin{figure*}[!t]
\includegraphics[width=\linewidth,height=1.55in]{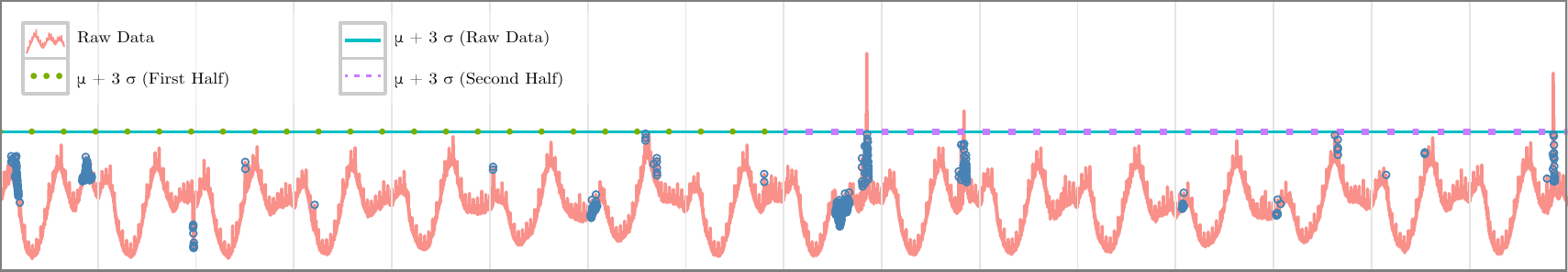}
\vspace{-3mm}
\caption{Application of $3 \cdot \sigma$ to the time series corresponding to \figref{fig:intro}}
\vspace{-4mm}
\label{fig:3sigma}
\end{figure*}

\section{Techniques} \label{sec:techniques}

In this section we detail our approaches {\bf Seasonal ESD (S-ESD)} and {\bf
Seasonal Hybrid ESD (S-H-ESD)}. Note that both the approaches are currently 
deployed to automatically detect anomalies in production data on a daily basis.
We employed an incremental approach in developing the two approaches. In 
particular, we started with evaluating the ``rule of thumb" (the {\em Three 
Sigma Rule}) using production data and progressively learned the limitations
of using the existing techniques for automatically detecting anomalies in 
production data. In the rest of this section we walk the reader through the
core steps of our aforementioned incremental approach. 

\subsection{Three-Sigma Rule} \label{sec:threesigma}

As a first cut, the $3 \cdot \sigma$ ``rule" is commonly used to detect anomalies 
in a given data set. Specifically, data points with values more than 3 times the
sample standard deviation are deemed anomalous. The rule can potentially be used
for capturing large global anomalies, but is ill-suited for detecting seasonal 
anomalies. This is exemplified by \figref{fig:3sigma}, wherein the seasonal anomalies 
that could not be captured using the $3 \cdot \sigma$ ``rule" are annotated with
circles. 

\begin{figure}[!h]
\centering
\includegraphics[width=1.0\linewidth]{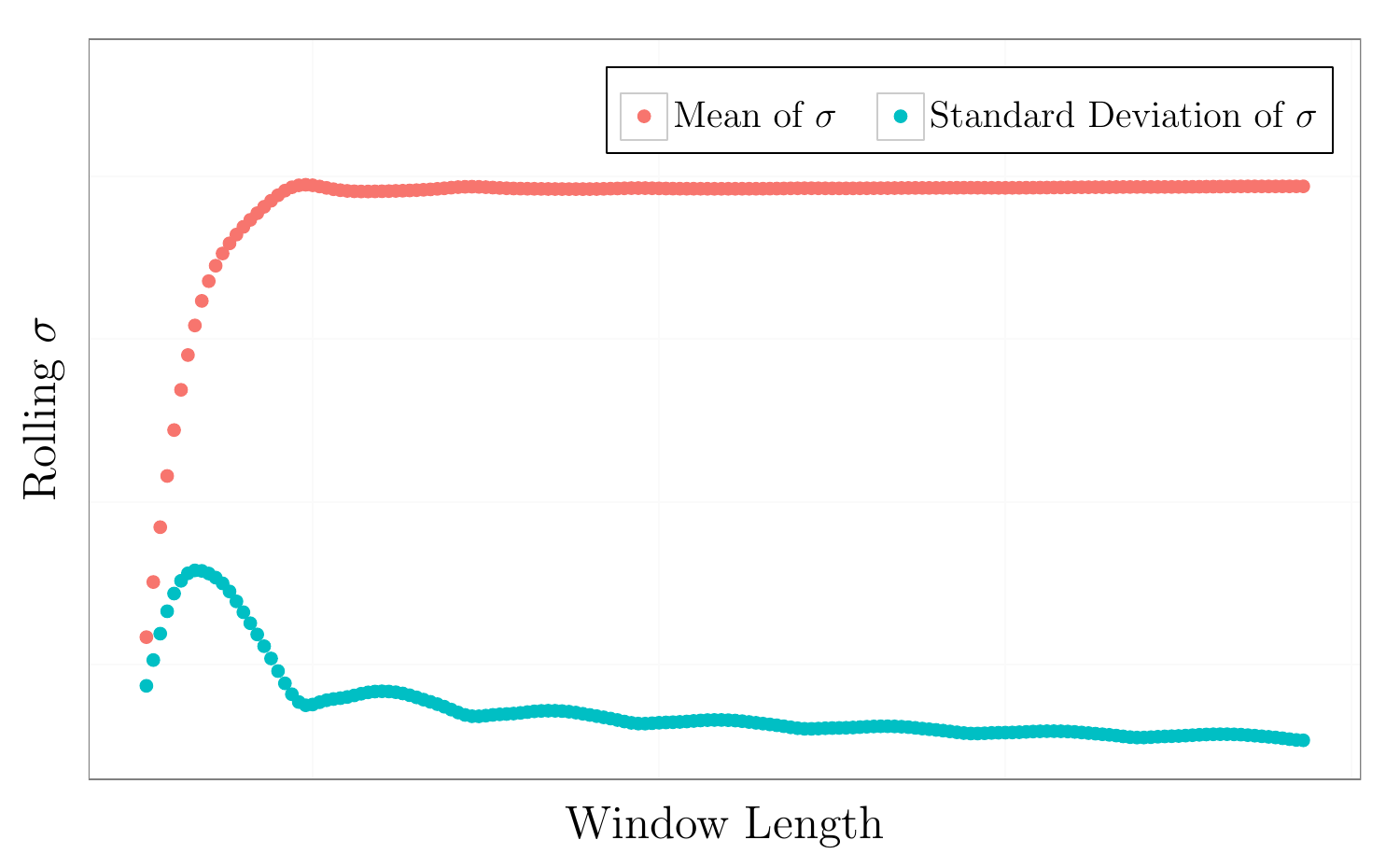}
\vspace{-2mm}
\caption{Distribution of $\sigma$ for different window lengths}
\vspace{-4mm}
\label{fig:var_sigma}
\end{figure}

Conceivably, one can potentially segment the input time series into multiple 
windows and apply the aforementioned rule on each window using their respective 
$\sigma$. The intuition behind segmentation being that the input time series is
non-stationary and consequently $\sigma$ varies over time. The variation in 
$\sigma$ for different window lengths is shown in \figref{fig:var_sigma}. 

\vspace{-2mm}

\noindent 
From \figref{fig:3sigma} we note that applying the rule on a per-window basis -- 
the time series was segmented into two windows -- does not facilitate capturing 
of the seasonal anomalies. 

\begin{figure}[!h] 
\centering 
\includegraphics[width=0.62\linewidth]{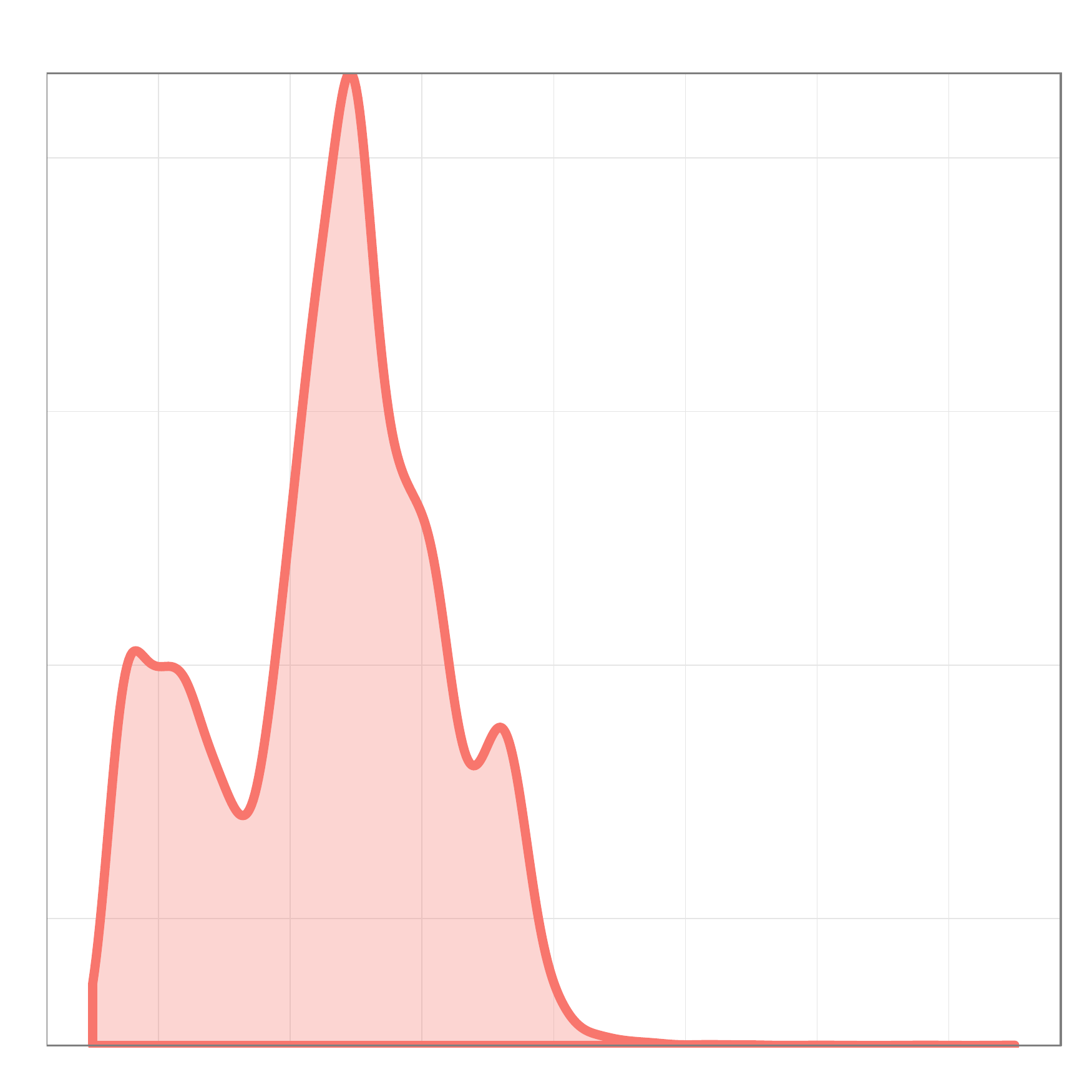}
\caption{Data distribution of the time series corresponding to \figref{fig:intro}}
\label{fig:multimodal} 
\end{figure}

The $3 \cdot \sigma$ ``rule" assumes that the underlying data distribution is 
normal. However, our experience has been that the time series data obtained from 
production is seldom, if at all, normal (see \figref{fig:multimodal} for an 
example).  

In light of the aforementioned limitations, we find that the $3 \cdot \sigma$
``rule" is not applicable in the current context. Next, we explored the efficacy
of using moving average for anomaly detection.

\begin{figure*}[!t]
        \begin{subfigure}{\linewidth}
                \includegraphics[width=\linewidth]{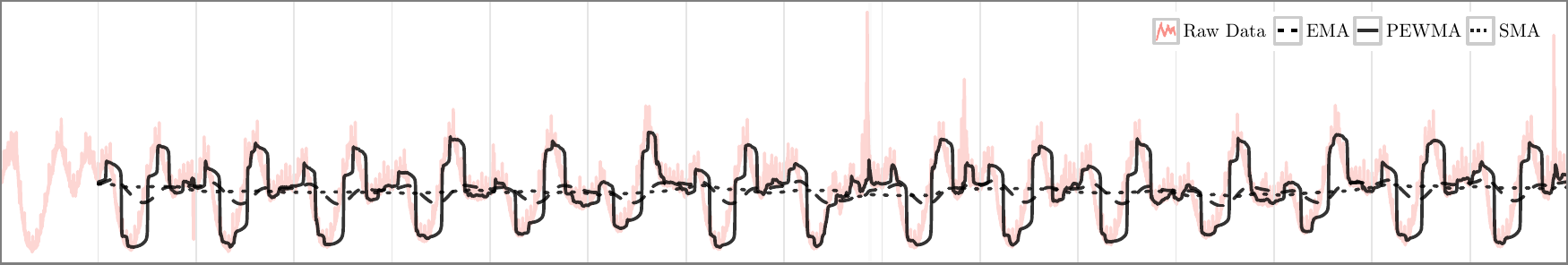}
                \caption{SMA, EWMA and PEWMA for the time series of \figref{fig:intro}}
                \label{fig:mov}
        \end{subfigure}
        ~ 
        \begin{subfigure}{\linewidth}
                \includegraphics[width=\linewidth]{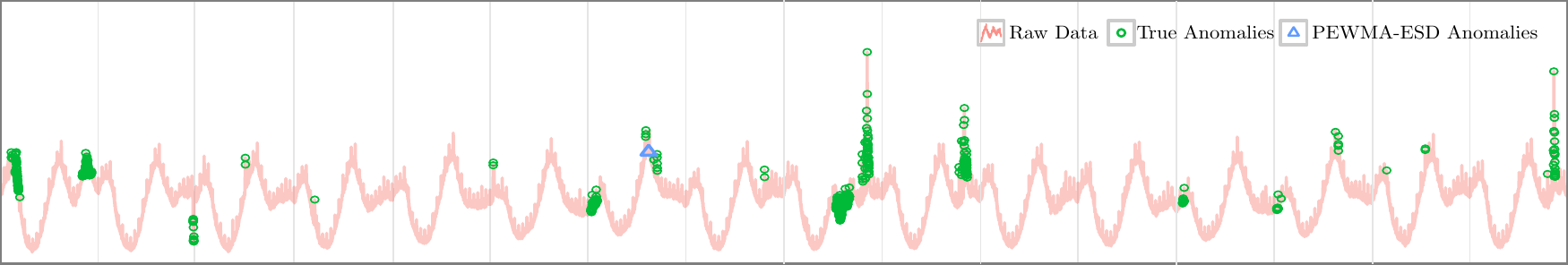}
                \caption{Scatter plot of ``true" anomalies and anomalies detected by applying ESD on PEWMA}
                \label{fig:movscatter}
        \end{subfigure}
\vspace{-2mm}
\caption{Illustration of limitations of using moving averages for anomaly detection}
\vspace{-2mm}
\label{fig:pewma}
\end{figure*}

\subsection{Moving Averages} \label{sec:movingaverages}

One of the key aspects associated with anomaly detection is to mitigate the 
impact of the presence of white noise. To this end, the use of moving averages has 
been proposed to filter (/smooth) out white noise. 

The most common moving average is the \textit{simple moving average} (SMA),
which is defined as: 

\vspace{-1mm}
\begin{equation} 
\text{SMA}_{t} = \frac{x_t + x_{t-1} + \ldots + x_{t-(n-1)}}{n} 
\end{equation}

Note that SMA weighs each of the previous $n$ data points equally. This may not 
desirable given the dynamic nature of the data stream observed in production and
from a recency perspective. To this end, the use of the {\em exponentially weighted 
moving average} (EWMA), defined by \eqnref{eq:ewma}, has been proposed \cite{
lucas_exponentially_1990}.

\vspace{-1mm}
\begin{equation} \text{EWMA}_{T}
\equiv \left\{
              \begin{matrix} 
                 \phantom{--------.}y_{t}=x_{t},&t=1\\ y_{t}=\alpha (x_{t}) + (1-\alpha)y_{t-1}, &t>1 
              \end{matrix}
       \right.  
\label{eq:ewma}
\end{equation}

\noindent
In \cite{carter_probabilistic_2012}, Carter and Streilein argue that in the context 
of streaming time series data EWMA can potentially be ``volatile to abrupt transient 
changes, losing utility for appropriately detecting anomalies''. To address this, 
they proposed the {\em Probabilistic Exponentially Weighted Moving Average} (PEWMA),
wherein the weighting parameter $\alpha$ is adapted by $(1 - \beta P_t)$, $P_t$ is 
the probability of $x_t$ evaluated under some modeled distribution, and $\beta$ is 
the weight placed on $P_t$. The SMA, EWMA and PEWMA for the time series of Figure  
\ref{fig:intro} is shown in \figref{fig:mov}. From the figure we note that the SMA 
and EWMA do not fare well with respect to capturing the daily seasonality. Although 
PEWMA traces the raw time series better (than SMA and EWMA), it fails to capture both 
global as well as seasonal anomalies. 

We evaluated the efficacy of using SMA, EWMA, and PEWMA as an input for anomaly 
detection. Our experience, using production data, was that the respective moving
averages filter out most of the seasonal anomalies and consequently are ill-suited 
for the current context. This is illustrated by \figref{fig:movscatter} from
which we note that application of ESD on PEWMA fails to detect ``true" anomalies
(annotated on the raw time series with circles). 

Further, given that a moving average is a lagging indicator, it is not suited for 
real time anomaly detection. 

Arguably, one can use shorter window length, when computing a moving average, to
better trace the input time series. However, as shown in \cite{Ahn03}, the standard
error of $\sigma$ ($\propto \frac{1}{\sqrt{n-1}}$), increases as the window length 
decreases. This is illustrated in \figref{fig:var_sigma} (window length decreases
from right to left).

\subsection{Seasonality and STL} \label{sec:stl}

As discussed in the previous section, Twitter data (obtained from production) 
exhibits heavy seasonality. Further, in most cases, the underlying distribution 
exhibits a multimodal distribution, as exemplified by \figref{fig:multimodal}. 
This limits the applicability of existing anomaly detection techniques such as 
Grubbs and ESD as the existing techniques assume a normal data distribution. 
Specifically, we learned based on data analysis that the presence of multiple
modes yields a higher value of standard deviation (the increase can be as much 
as up to 5\%) which in turn leads to masking of detection of some of the ``true"
anomalies. 

To this end, we employ time series decomposition wherein a given time series
($X$) is decomposed into three -- seasonal ($S_X$), trend ($T_X$), and residual 
($R_X$) -- components. The residual has a unimodal distribution that is amenable 
to the application of anomaly detection techniques such as ESD. Before moving on,
let us first define {\em sub-cycle series} \cite{cleveland_stl:_1990}: 

\begin{definition}
A sub-cycle series comprises of values at each position of a seasonal cycle. 
For example, if the series is monthly with a yearly periodicity, then the 
first sub-cycle series is the January values, the second is the February 
values, and so forth. 
\end{definition}

Time series decomposition approaches can be either additive or multiplicative 
\cite{stuart_advanced_1983}, with additive decomposition being appropriate for 
seasonal data that has a constant magnitude as is the case in the current context. 
The algorithm first derives $T_X$ using a moving average filter, and subsequently 
subtracts it from the $X$. Once the trend is removed, $S_X$ is then estimated as 
the average of the data points within the corresponding sub-cycle series. Note 
that the number of data points per period is an input to the algorithm in order 
to create the sub-cycle series. In the current context, we set the number of data 
points per period to be a function of the data granularity. The residual $R_X$ 
is estimated as follows:

\vspace{-1mm}
\begin{equation} 
R_X = X - T_X - S_X 
\label{eq:res}
\end{equation}

\noindent 
The residual component computed above can be potentially corrupted by extreme 
anomalies in the input data. 

\begin{figure*}[!t]
\centering
\includegraphics[width=\textwidth]{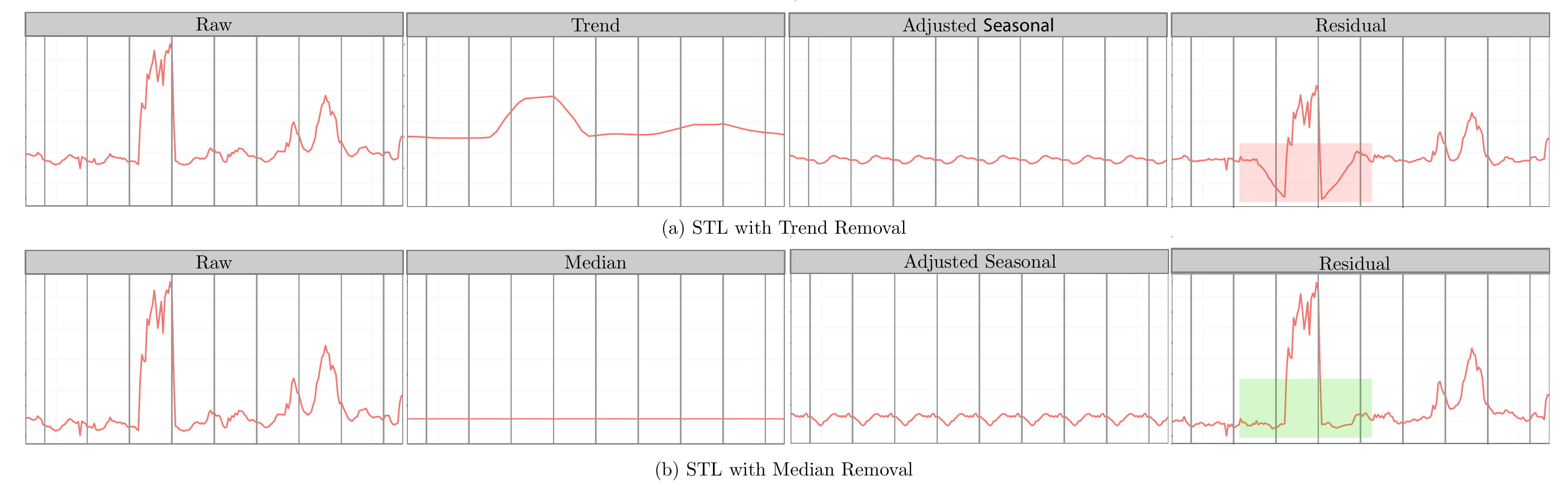}
\vspace*{-2mm}
\caption{STL (a) vs. STL Variant (b) Decomposition}
\vspace*{-2mm}
\label{fig:stlvsstlmedian}
\end{figure*}

This issue is addressed by STL \cite{cleveland_stl:_1990}, a robust approach 
to decomposition that uses LOESS\cite{cleveland_robust_1979} to estimate the
seasonal component. The algorithm consists of an inner loop that derives the
trend, seasonal, and residual components and an outer loop that increases the
robustness of the algorithm with respect to anomalies. Akin to classical 
seasonal decomposition, the inner loop derives the trend component using a 
moving average filter, removes the trend from the data and then smoothes the 
sub-cycle series to derive the seasonal component; however, STL uses LOESS
to derive the seasonality, allowing the decomposition to fit more complex
functions than either the additive or multiplicative approaches. Additionally,
STL iteratively converges on the decomposition, repeating the process several
times, or until the difference in iterations is smaller then some specified
threshold ($\epsilon$). Optionally, STL can be made more robust against the 
influence of anomalies by weighting data points in an outer loop. The outer 
loop uses the decomposed components to assign a robustness weight to each 
datum as:

\begin{equation*} 
Weight_{t}=B\left(\frac{|R_{X_t}|}{6\times median|R_{X_t}|}\right)
\end{equation*}

\noindent 
where B is the bisquare function defined as:

\begin{equation*} 
B(u) \equiv \left\{\begin{matrix} (1 - u^{2})^{2}&for\phantom{-}u\leq 0< 1\\ 0&for\phantom{-}u>1\phantom{.......} \end{matrix}\right.
\end{equation*}

\noindent 
These weights are then used to converge closer to the ``true" decomposed 
components in the next iteration of the inner loop.

\begin{figure*}[!t] 
\centering
\includegraphics[width=\textwidth]{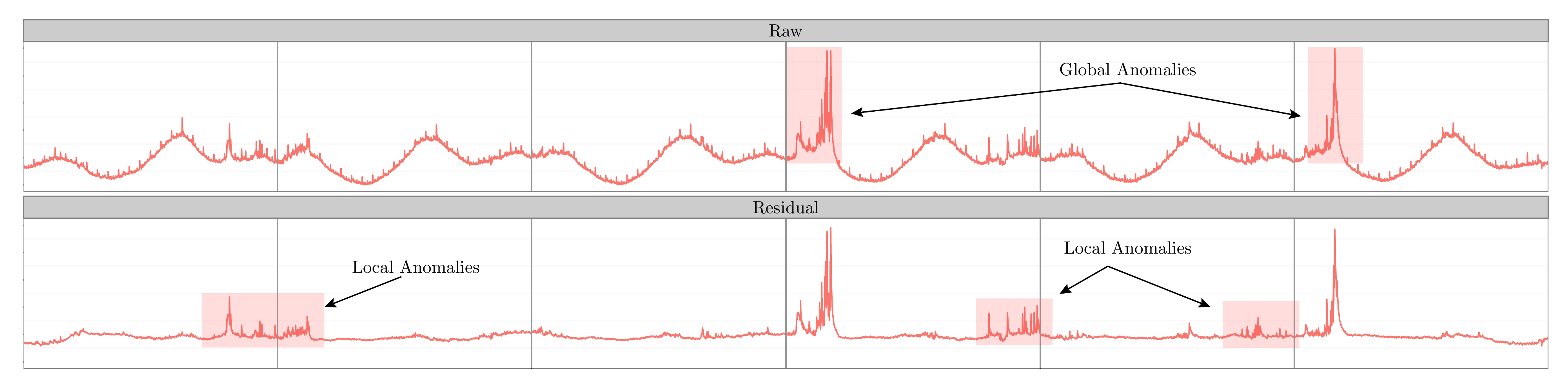}
\caption{Global and Local Anomalies Exposed using S-ESD}
\label{fig:globallocalanoms} 
\end{figure*}

\subsection{Seasonal ESD (S-ESD)} \label{sec:sesd}

To recap, the applicability of the existing techniques (overviewed in Section 
\ref{sec:background}) is limited by the following: 

\begin{dinglist}{122}
\item Presence of seasonality in the Twitter time series data (exemplified by 
      \figref{fig:intro}). 
\item Multimodal data distribution of Twitter time series data (illustrated
      by \figref{fig:multimodal}). 
\end{dinglist}

\noindent 
To this end, we propose a novel algorithm, referred to as {\em Seasonal-ESD} 
(S-ESD), to automatically detect anomalies in Twitter's production data. The 
algorithm uses a modified STL decomposition (discussed in \sssecref{stl_var})
to extract the residual component of the input time series and then applies 
ESD to detect anomalies. 

This two step process allows S-ESD to detect \underline{both} global anomalies
that extend beyond the expected seasonal minimum and maximum and local anomalies 
that would otherwise be masked by the seasonality. 

A formal description of the algorithm is presented in \algoref{algo:s-esd-algo}. 

In the rest of this subsection, we detail the modified STL algorithm, elaborate
S-ESD's ability to detect global and local anomalies, as well as the limitations 
of the algorithm. 

\secref{sec:eval} presents the efficacy of S-ESD using production data -- both
core drivers (or business metrics) and system metrics. 

\begin{algorithm}[!h]
\begin{algorithmic}
\STATE \ \
\STATE {\bf Input}: \\[1mm]
 	$X$ = A time series \\[1mm]
 	$n$ = number of observations in $X$ \\[1mm]
 	$k$ = max anomalies (iterations in ESD)\\
\bigskip
\STATE {\bf Output:} \\[1mm]
 	$X_A$ = An anomaly vector wherein each element is a tuple $(timestamp, observed \; value)$
\bigskip
\STATE {\bf Require:} \\[1mm]
 		$k \leq (n \times .49)$ \\[1mm]
\STATE 1. Extract seasonal component $S_X$ using STL Variant \\[1mm] 
\STATE 2. Compute median $\tilde{X}$ \\[1mm]
\STATE /* Compute residual */ \\[1mm]
\STATE 3. $R_X = X - S_X - \tilde{X}$ \\[1mm]
\STATE /* Detect anomalies vector $X_A$ using ESD */ \\[1mm]
\STATE 4. $X_A$ = ESD($R$, $k$) \\
\bigskip
\RETURN $X_A$

\end{algorithmic}
\caption{S-ESD Algorithm}
\label{algo:s-esd-algo}
\end{algorithm}

\subsubsection{STL Variant} \label{stl_var} 

Applying STL decomposition to time series of system metrics yielded, in some 
cases, spurious anomalies (i.e., anomalies not present in the original time 
series) in the residual component. For example, let us consider the time 
series shown in \figref{fig:stlvsstlmedian}a wherein we observe a region 
of continuous anomalies in the raw data. On applying STL decomposition, we
observed a breakout (ala a pulse) in the trend component. On deriving the 
residual component using \eqnref{eq:res}, we observed an inverse breakout,
highlighted with a red rectangle in \figref{fig:stlvsstlmedian}a, which 
in turn yielded spurious anomalies.

To address the above, we use the median of the time series to represent the 
``stable'' trend value which is in turn used to compute the residual component 
as follows:  

\begin{equation} 
R_X = X - S_X - \tilde{X} 
\end{equation}

\noindent
where $X$ is the raw time series, $S_X$ is the seasonal component as determined
by STL, and $\tilde{X}$ is the median of the raw time series. Replacing the trend 
with the median eliminates the spurious anomalies in the residual component as
exemplified by \figref{fig:stlvsstlmedian}b. From the figure we note that the
region highlighted with a green rectangle does not have any spurious anomalies,
unlike the corresponding region in \figref{fig:stlvsstlmedian}a.

\begin{figure*}[!t]
\centering
\begin{subfigure}{\linewidth}
  \includegraphics[width=\linewidth]{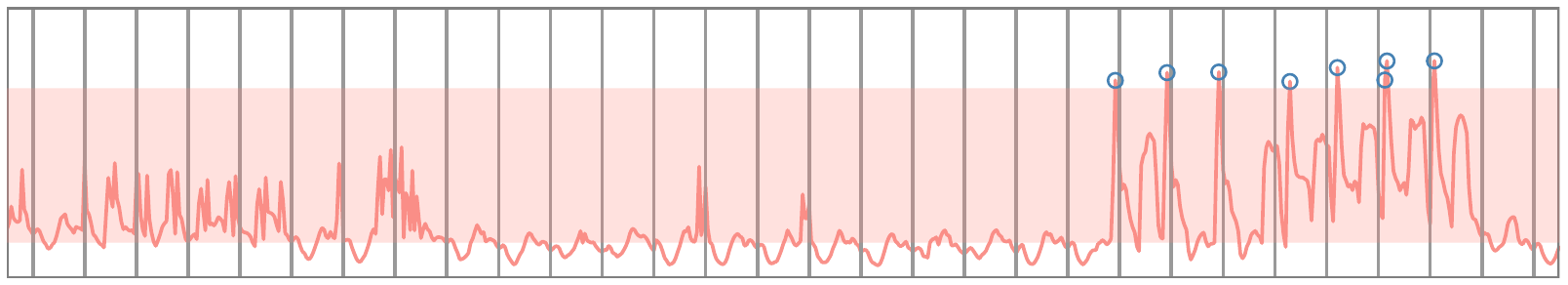}
  \caption{Anomalies detected via S-ESD: 1.11\% Anomalies ($\alpha = $0.05)} 
  \label{fig:s-esd-prod-blobstore} 
\end{subfigure} 
\begin{subfigure}{\linewidth}
  \includegraphics[width=\linewidth]{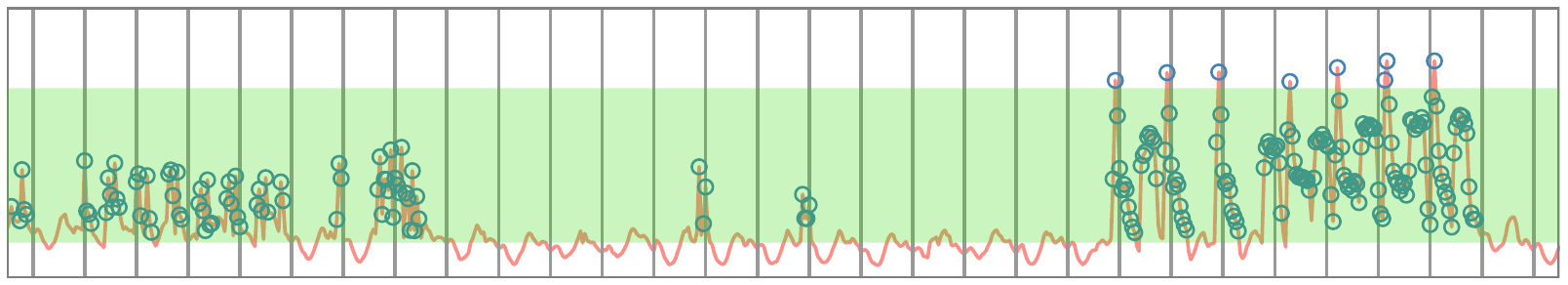}
  \caption{Anomalies detected via S-H-ESD: 29.68\% Anomalies ($\alpha = $0.05)} 
  \label{fig:s-h-esd-prod-blobstore} 
\end{subfigure}
\caption{S-ESD (a) vs. S-H-ESD performance with highly anomalous data}
\label{fig:esd_comparison} 
\end{figure*}

\subsubsection{Global and Local Anomalies} \label{sec:decompglobloc} 

Unlike the techniques overviewed in \secref{sec:background}, S-ESD can detect
local anomalies that would otherwise be masked by seasonal data. These local
anomalies are bound between the seasonal minimum and maximum and may not not
appear to be anomalous from a global perspective. However, they are indeed
anomalous and it is important to detect these as they represent a deviation 
from the historical pattern. For instance, local anomalies found in Twitter 
data and other social network data, may reflect changes in user behavior, or the
systems in the data center/cloud. \figref{fig:globallocalanoms} illustrates
the use of STL variant to expose both global and local anomalies.

\subsubsection{S-ESD Limitations} \label{sec:sesd_limits} 

Although S-ESD can be used for detection of both global and local anomalies, 
S-ESD does not fare well when applied to data sets that have a high percentage 
of anomalies. This is exemplified by \figref{fig:s-esd-prod-blobstore} wherein
S-ESD does not capture the anomalies corresponding to the region highlighted 
by the red rectangle. 

As discussed earlier in \ssecref{sec:robuststats}, a single large value can 
inflate both the mean and the standard deviation. This makes ESD conservative 
in tagging anomalies and results in a large number of false negatives. In the
following section we detail the application of robust statistics as a further
refinement of S-ESD.

\subsection{Seasonal Hybrid ESD (S-H-ESD)} \label{sec:shesd}

Seasonal Hybrid ESD (S-H-ESD) builds upon the S-ESD algorithm described in
the previous subsection. In particular, S-H-ESD uses the robust statistical 
techniques and metrics discussed in \ssecref{sec:robuststats} to enable a 
more consistent measure of central tendency of a time series with a high 
percentage of anomalies. For example, let us consider the time series shown 
in \figref{fig:s-esd-prod-blobstore}. From the graph we observe that the 
seasonal component is apparent in the middle region of the time series; 
however, a significantly large portion of the time series is anomalous. This 
can inflate the mean and standard deviation, resulting in true anomalies
being mislabeled as not anomalous and consequently yielding a high 
number of false negatives.

\begin{table}[!h] 
\centering 
\begin{tabular}{l|l|l|l} 
\textbf{Mean}     & \textbf{Median}    & \textbf{Std. Dev.} & \textbf{MAD} \\ \hline \hline
6.59 & 5.45 & 3.08 & 1.52 \\ \hline \hline
\end{tabular}
\caption{Comparison of Mean vs. Median and Standard Deviation vs. Median Absolute Deviation} 
\label{tab:shesdtable} 
\end{table}

\noindent 
We addressed the above by replacing the mean and standard deviation used in
ESD with more robust statistical measures during the calculation of the test
statistic (refer to \eqnref{eq:esd-test-stat}); in particular, we use the 
median and MAD, as these metrics exhibit a higher breakdown point (discussed 
earlier in \secref{sec:robuststats}). 

\tabref{tab:shesdtable} lists the aforementioned metrics for the time series
in \figref{fig:esd_comparison}. The anomalies induce a small difference between
the mean and median ($\approx$ 1.2\%), however the standard deviation is 
$> 2\times$ the median absolute deviation. This results in S-ESD detecting 
only 1.11\% of the data as ``anomalous'', whereas S-H-ESD correctly detects 
29.68\% of the input time series as anomalous -- contrast the two graphs 
shown in \figref{fig:s-h-esd-prod-blobstore}. 

Note that using the median and MAD requires sorting the data and consequently
the run time of S-H-ESD is higher than that of S-ESD. Therefore, in cases where
the time series under consideration is large but with a relatively low anomaly
count, it is advisable to use S-ESD. A detailed performance comparison between
the two approaches is presented in \secref{sec:efficacy}.

\begin{table*}[ht]
\centering
\resizebox{\linewidth}{!}{
\begin{tabular}{|c|l|l|l|l|l|l|l|l|l|l|l|l|l|}
\hline
\textit{One-Tail (Capacity)}                      & \multicolumn{6}{|c|}{\textit{Alpha = 0.05}}                                                                                                                                                                                                             & \multicolumn{6}{|c|}{\textit{Alpha = 0.001}}                                                                                                                                                                                                            & \multicolumn{1}{|c|}{\# of Observations} \\ \hline
\multicolumn{1}{|l|}{}                            & \multicolumn{3}{|c|}{\textbf{S-ESD}}                                                                                       & \multicolumn{3}{|c|}{\textbf{S-H-ESD}}                                                                                     & \multicolumn{3}{|c|}{\textbf{S-ESD}}                                                                                       & \multicolumn{3}{|c|}{\textbf{S-H-ESD}}                                                                                     &                                          \\ \hline
\textbf{Dataset \#}                       & \textit{Precision}                       & \textit{Recall}                       & \textit{Fmeasure}                       & \textit{Precision}                       & \textit{Recall}                       & \textit{Fmeasure}                       & \textit{Precision}                       & \textit{Recall}                       & \textit{Fmeasure}                       & \textit{Precision}                       & \textit{Recall}                       & \textit{Fmeasure}                       &                                          \\ \hline
System 1                                  & 1.00                                     & 0.05                                  & 0.09                                    & 0.26                                     & 1.00                                  & 0.41                                    & 1.00                                     & 0.02                                  & 0.05                                    & 0.28                                     & 1.00                                  & 0.44                                    & 337                                      \\ \hline
System 2									 & 1.00                                     & 0.04                                  & 0.07                                    & 0.95                                     & 0.87                                  & 0.91                                    & 1.00                                     & 0.00                                  & 0.01                                    & 0.95                                     & 0.81                                  & 0.88                                    & 721                                      \\ \hline
System 3									 & 1.00                                     & 0.14                                  & 0.25                                    & 0.98                                     & 0.93                                  & 0.95                                    & 1.00                                     & 0.01                                  & 0.01                                    & 1.00                                     & 0.77                                  & 0.87                                    & 601                                      \\ \hline
System 4									 & 1.00                                     & 0.57                                  & 0.73                                    & 0.71                                     & 0.99                                  & 0.82                                    & 1.00                                     & 0.43                                  & 0.60                                    & 0.84                                     & 0.95                                  & 0.89                                    & 913                                      \\ \hline
System 5									 & 0.98                                     & 1.00                                  & 0.99                                    & 0.98                                     & 1.00                                  & 0.99                                    & 0.98                                     & 1.00                                  & 0.99                                    & 0.98                                     & 1.00                                  & 0.99                                    & 337                                      \\ \hline
System 6									 & 0.75                                     & 0.28                                  & 0.41                                    & 0.65                                     & 0.34                                  & 0.45                                    & 0.80                                     & 0.13                                  & 0.22                                    & 0.75                                     & 0.28                                  & 0.41                                    & 721                                      \\ \hline
\multicolumn{1}{|c|}{\textbf{Average}}   & \BlackCell{0.96}                         & \BlackCell{0.35}                          & \BlackCell{0.42}                        & \BlackCell{0.75} 						   & \BlackCell{0.85}                                  & \BlackCell{0.76}                      	 & \BlackCell{0.96}                         & \BlackCell{0.26}                                  & \BlackCell{0.31}                        & \BlackCell{0.80}                         & \BlackCell{0.80}                                  & \BlackCell{0.75}                        &                                          \\ \hline
Core Driver 1                             & 0.99                                     & 0.48                                  & 0.65                                    & 0.83                                     & 0.78                                  & 0.80                                    & 1.00                                     & 0.40                                  & 0.57                                    & 0.95                                     & 0.61                                  & 0.74                                    & 43195                                    \\ \hline
Core Driver 2                             & 1.00                                     & 0.03                                  & 0.06                                    & 0.96                                     & 0.25                                  & 0.39                                    & 1.00                                     & 0.01                                  & 0.02                                    & 1.00                                     & 0.05                                  & 0.10                                    & 43200                                    \\ \hline
Core Driver 3                             & 0.08                                     & 1.00                                  & 0.14                                    & 0.05                                     & 1.00                                  & 0.10                                    & 0.12                                     & 1.00                                  & 0.22                                    & 0.08                                     & 1.00                                  & 0.14                                    & 43200                                    \\ \hline
Core Driver 4                             & 0.34                                     & 0.20                                  & 0.25                                    & 0.20                                     & 0.30                                  & 0.24                                    & 0.54                                     & 0.18                                  & 0.27                                    & 0.29                                     & 0.22                                  & 0.25                                    & 43200                                    \\ \hline
Core Driver 5                             & 0.36                                     & 0.98                                  & 0.52                                    & 0.19                                     & 0.98                                  & 0.32                                    & 0.53                                     & 0.98                                  & 0.69                                    & 0.33                                     & 0.98                                  & 0.49                                    & 43200                                    \\ \hline
Core Driver 6                             & 0.08                                     & 0.69                                  & 0.14                                    & 0.06                                     & 0.69                                  & 0.11                                    & 0.12                                     & 0.69                                  & 0.20                                    & 0.08                                     & 0.69                                  & 0.14                                    & 43200                                    \\ \hline
Core Driver 7                             & 0.27                                     & 0.97                                  & 0.42                                    & 0.20                                     & 0.97                                  & 0.33                                    & 0.33                                     & 0.97                                  & 0.49                                    & 0.25                                     & 0.97                                  & 0.40                                    & 43200                                    \\ \hline
Core Driver 8                             & 0.66                                     & 0.95                                  & 0.78                                    & 0.58                                     & 0.99                                  & 0.73                                    & 0.80                                     & 0.63                                  & 0.71                                    & 0.67                                     & 0.95                                  & 0.78                                    & 43200                                    \\ \hline
Core Driver 9                             & 0.80                                     & 0.54                                  & 0.65                                    & 0.58                                     & 0.63                                  & 0.60                                    & 0.91                                     & 0.53                                  & 0.67                                    & 0.72                                     & 0.56                                  & 0.63                                    & 43178                                    \\ \hline
Core Driver 10                            & 0.82                                     & 0.20                                  & 0.32                                    & 0.67                                     & 0.30                                  & 0.42                                    & 1.00                                     & 0.08                                  & 0.14                                    & 0.81                                     & 0.20                                  & 0.32                                    & 43200                                    \\ \hline
\multicolumn{1}{|c|}{\textbf{Average}}   & \BlackCell{0.54}                         & \BlackCell{0.60}                                  & \BlackCell{0.39}                        & \BlackCell{0.43}                         & \BlackCell{0.69}                                  & \BlackCell{0.40}                        & \BlackCell{0.63}                         & \BlackCell{0.55}                          & \BlackCell{0.40}                        & \BlackCell{0.52}                         & \BlackCell{0.62}                                  & \BlackCell{0.40}                        &                                          \\ \hline
\end{tabular}
}
\caption{CapEng Perspective: Precision, Recall, and F-measure} 
\label{tab:capeng}
\end{table*}

\begin{table*}[ht]
\centering
\resizebox{\linewidth}{!}{
\begin{tabular}{|c|l|l|l|l|l|l|l|l|l|l|l|l|l|}
\hline
One-Tail (User Behavior)                  & \multicolumn{6}{|c|}{\textit{Alpha = 0.05}}                                                                                                                                                                                                             & \multicolumn{6}{|c|}{\textit{Alpha = 0.001}}                                                                                                                                                                                                            & \multicolumn{1}{|c|}{\# of Observations} \\ \hline
\multicolumn{1}{|l|}{}                    & \multicolumn{3}{|c|}{\textbf{S-ESD}}                                                                                       & \multicolumn{3}{|c|}{\textbf{S-H-ESD}}                                                                                     & \multicolumn{3}{|c|}{\textbf{S-ESD}}                                                                                       & \multicolumn{3}{|c|}{\textbf{S-H-ESD}}                                                                                     &                                          \\ \hline
\textbf{Dataset \#}                               & \multicolumn{1}{|c|}{\textit{Precision}} & \multicolumn{1}{|c|}{\textit{Recall}} & \multicolumn{1}{|c|}{\textit{Fmeasure}} & \multicolumn{1}{|c|}{\textit{Precision}} & \multicolumn{1}{|c|}{\textit{Recall}} & \multicolumn{1}{|c|}{\textit{Fmeasure}} & \multicolumn{1}{|c|}{\textit{Precision}} & \multicolumn{1}{|c|}{\textit{Recall}} & \multicolumn{1}{|c|}{\textit{Fmeasure}} & \multicolumn{1}{|c|}{\textit{Precision}} & \multicolumn{1}{|c|}{\textit{Recall}} & \multicolumn{1}{|c|}{\textit{Fmeasure}} &                                          \\ \hline
System 1                                  & 1.00                                     & 0.02                                  & 0.04                                    & 0.60                                     & 1.00                                  & 0.75                                    & 1.00                                     & 0.01                                  & 0.02                                    & 0.65                                     & 1.00                                  & 0.79                                    & 337                                      \\ \hline
System 2                                  & 1.00                                     & 0.03                                  & 0.06                                    & 1.00                                     & 0.68                                  & 0.81                                    & 1.00                                     & 0.00                                  & 0.01                                    & 1.00                                     & 0.63                                  & 0.77                                    & 721                                      \\ \hline
System 3                                  & 1.00                                     & 0.12                                  & 0.22                                    & 1.00                                     & 0.82                                  & 0.90                                    & 1.00                                     & 0.00                                  & 0.01                                    & 1.00                                     & 0.67                                  & 0.80                                    & 601                                      \\ \hline
System 4                                  & 1.00                                     & 0.49                                  & 0.66                                    & 0.84                                     & 1.00                                  & 0.91                                    & 1.00                                     & 0.37                                  & 0.54                                    & 1.00                                     & 0.96                                  & 0.98                                    & 913                                      \\ \hline
System 5                                  & 0.99                                     & 1.00                                  & 0.99                                    & 0.99                                     & 1.00                                  & 0.99                                    & 0.99                                     & 1.00                                  & 0.99                                    & 0.99                                     & 1.00                                  & 0.99                                    & 337                                      \\ \hline
System 6                                  & 1.00                                     & 0.38                                  & 0.55                                    & 1.00                                     & 0.53                                  & 0.69                                    & 1.00                                     & 0.16                                  & 0.27                                    & 1.00                                     & 0.38                                  & 0.55                                    & 721                                      \\ \hline
\multicolumn{1}{|c|}{\textbf{Average}}   & \BlackCell{1.00}                         & \BlackCell{0.34}                                  & \BlackCell{0.42}                        & \BlackCell{0.90}                         & \BlackCell{0.84}                                  & \BlackCell{0.84}                        & \BlackCell{1.00}                         & \BlackCell{0.26}                                  & \BlackCell{0.31}                        & \BlackCell{0.94}                         & \BlackCell{0.77}                                  & \BlackCell{0.81}                        &                                          \\ \hline
Core Driver 1                             & 1.00                                     & 0.46                                  & 0.63                                    & 1.00                                     & 0.90                                  & 0.95                                    & 1.00                                     & 0.38                                  & 0.55                                    & 1.00                                     & 0.61                                  & 0.76                                    & 43195                                    \\ \hline
Core Driver 2                             & 1.00                                     & 0.03                                  & 0.05                                    & 1.00                                     & 0.21                                  & 0.35                                    & 1.00                                     & 0.01                                  & 0.02                                    & 1.00                                     & 0.04                                  & 0.08                                    & 43200                                    \\ \hline
Core Driver 3                             & 1.00                                     & 0.27                                  & 0.43                                    & 1.00                                     & 0.40                                  & 0.57                                    & 1.00                                     & 0.18                                  & 0.30                                    & 1.00                                     & 0.28                                  & 0.43                                    & 43200                                    \\ \hline
Core Driver 4                             & 1.00                                     & 0.03                                  & 0.05                                    & 1.00                                     & 0.07                                  & 0.13                                    & 1.00                                     & 0.02                                  & 0.03                                    & 1.00                                     & 0.03                                  & 0.07                                    & 43200                                    \\ \hline
Core Driver 5                             & 1.00                                     & 0.22                                  & 0.36                                    & 1.00                                     & 0.41                                  & 0.58                                    & 1.00                                     & 0.15                                  & 0.25                                    & 1.00                                     & 0.23                                  & 0.38                                    & 43200                                    \\ \hline
Core Driver 6                             & 1.00                                     & 0.39                                  & 0.56                                    & 1.00                                     & 0.52                                  & 0.68                                    & 1.00                                     & 0.27                                  & 0.42                                    & 1.00                                     & 0.39                                  & 0.56                                    & 43200                                    \\ \hline
Core Driver 7                             & 1.00                                     & 0.72                                  & 0.84                                    & 1.00                                     & 0.96                                  & 0.98                                    & 1.00                                     & 0.59                                  & 0.74                                    & 1.00                                     & 0.77                                  & 0.87                                    & 43200                                    \\ \hline
Core Driver 8                             & 1.00                                     & 0.24                                  & 0.39                                    & 1.00                                     & 0.29                                  & 0.45                                    & 1.00                                     & 0.13                                  & 0.24                                    & 1.00                                     & 0.24                                  & 0.39                                    & 43200                                    \\ \hline
Core Driver 9                             & 1.00                                     & 0.35                                  & 0.52                                    & 1.00                                     & 0.56                                  & 0.72                                    & 1.00                                     & 0.30                                  & 0.46                                    & 1.00                                     & 0.40                                  & 0.58                                    & 43178                                    \\ \hline
Core Driver 10                            & 1.00                                     & 0.14                                  & 0.25                                    & 1.00                                     & 0.28                                  & 0.43                                    & 1.00                                     & 0.05                                  & 0.09                                    & 1.00                                     & 0.15                                  & 0.26                                    & 43200                                    \\ \hline
\multicolumn{1}{|c|}{\textbf{Average}}   & \BlackCell{1.00}                         & \BlackCell{0.29}                                  & \BlackCell{0.41}                        & \BlackCell{1.00}                         & \BlackCell{0.46}                                  & \BlackCell{0.58}                        & \BlackCell{1.00}                         & \BlackCell{0.21}                                  & \BlackCell{0.31}                        & \BlackCell{1.00}                         & \BlackCell{0.32}                                  & \BlackCell{0.44}                        &                                          \\ \hline
\end{tabular}
}
\caption{User Behavior Perspective: Precision, Recall, and F-measure} 
\label{tab:ub}
\end{table*}

\section{Evaluation} \label{sec:eval}

In this section we outline our methodology for evaluating the proposed 
techniques, discuss the deployment in production, and last but not the 
least, present results to demonstrate the efficacy of the proposed 
techniques. 

\subsection{Methodology} \label{sec:methodology}

The efficacy of S-ESD and S-H-ESD was evaluated using a wide corpus of time 
series data obtained from \underline{production}. The time series corresponded
to both low-level {\em system metrics} and higher-level {\em core drivers} 
(business metrics). For example, but not limited to, the following metrics 
were used:

\begin{dinglist}{114} 
\item System Metrics 
\begin{dinglist}{122}
  \item CPU utilization 
  \item Heap usage
  \item Time spent in GC (garbage collection)
  \item Disk writes
\end{dinglist}

\item Application Metrics
\begin{dinglist}{122} 
  \item Request rate
  \item Latency
\end{dinglist}

\item Core Drivers 
\begin{dinglist}{122} 
  \item Tweets per minute (TPM) 
  \item Retweets per minute (RTPM) 
  \item Unique Photos Per Minute (UPPM) 
\end{dinglist}
\end{dinglist}

\noindent 
More than 20 data sets were used for evaluation. The system metrics ranged from
two-week long periods to four-week long periods, with hour granularity. The core 
drivers metrics were all four-week periods, with minute granularity.

\subsection{Production} \label{sec:production}

Increasingly, machine-generated BigData is being used to drive performance 
and efficiency of data centers/cloud computing platforms. BigData is often 
characterized by \textit{volume} and \textit{velocity} \cite{BigData9,BigData10}. Given the 
multitude of services in our service-oriented-architecture (SOA), and the 
fact that each service monitors a large set of metrics, it is imperative 
to {\em automatically} detect anomalies\footnote{A manual approach would
be prohibitive from a cost perspective and would also be error-prone.} in 
the time series of each metric. We have deployed the proposed techniques 
for automatic detection of anomalies in production data for a wide set of
services. 

One can also set a threshold to refine the set of anomalies detected (using
the proposed techniques) based on the specific requirements of the 
service owner. For example, it is possible for capacity engineers to set a 
threshold such that only anomalies greater than the specified threshold are 
reported.

Based on extensive experimentation and analysis, S-H-ESD with $\alpha=0.05$
(95\% confidence) was selected for detecting anomalies in the metrics mentioned 
in the previous subsection. For each metric, S-H-ESD is run over a time series 
containing the last 14 days worth of data and an e-mail report is sent out {\em if} 
one or more anomalies were detected the previous day. The anomalies and time series 
are plotted using an in-house data visualization and analytics framework called Chiffchaff. 
Chiffchaff uses the ggplot2 plotting environment in R to produce graphs similar
to the graphs shown in this paper. Additionally, CSVs with the metric, timestamps
and magnitude of any anomalies detected is also attached to the email report.

\begin{table*}
\centering
\resizebox{\linewidth}{!}{
\begin{tabular}{|c|c|c|c|c|c|c|c|c|c|c|c|c|c|}
\hline
\textit{One-Tail (Injection)}                                               & \multicolumn{6}{|c|}{\textit{Alpha = 0.05}}                                                                         & \multicolumn{6}{|c|}{\textit{Alpha = 0.001}}                                                                        & \textit{\# of Observations} \\ \hline
                                                                & \multicolumn{3}{|c|}{\textbf{S-ESD}}                     & \multicolumn{3}{|c|}{\textbf{S-H-ESD}}                   & \multicolumn{3}{|c|}{\textbf{S-ESD}}                     & \multicolumn{3}{|c|}{\textbf{S-H-ESD}}                   &                             \\ \hline
\textbf{Dataset}                                                & \textit{Precision} & \textit{Recall} & \textit{Fmeasure} & \textit{Precision} & \textit{Recall} & \textit{Fmeasure} & \textit{Precision} & \textit{Recall} & \textit{Fmeasure} & \textit{Precision} & \textit{Recall} & \textit{Fmeasure} &                             \\ \hline
mag.75\_width5\_pct100 & 1                  & 0.72       & 0.84          & 1                  & 0.79       & 0.88         & 1                  & 0.61       & 0.76         & 1                  & 0.70       & 0.82         & 43200                       \\ \hline
mag1.5\_width5\_pct100                                          & 1                  & 0.96        & 0.98         & 1                  & 0.99       & 0.99         & 1                  & 0.89       & 0.94         & 1                  & 0.94       & 0.97         & 43200                       \\ \hline
mag3\_width5\_pct100                                            & 1                  & 1               & 1                 & 1                  & 1               & 1                 & 1                  & 1               & 1                 & 1                  & 1               & 1                 & 43200                       \\ \hline
mag3\_width10\_pct100 & 1                  & 1               & 1                 & 1                  & 1               & 1                 & 1                  & 1               & 1                 & 1                  & 1               & 1                 & 43200                       \\ \hline
mag3\_width25\_pct100                                           & 1                  & 1               & 1                 & 1                  & 1               & 1                 & 1                  & 1               & 1                 & 1                  & 1               & 1                 & 43200                       \\ \hline
mag3\_width50\_pct100                                           & 1                  & 1               & 1                 & 1                  & 1               & 1                 & 1                  & 1               & 1                 & 1                  & 1               & 1                 & 43200                       \\ \hline
mag3\_width100\_pct100                                          & 1                  & 1               & 1                 & 1                  & 1               & 1                 & 1                  & 1               & 1                 & 1                  & 1               & 1                 & 43200                       \\ \hline
mag6\_width5\_pct100                                            & 1                  & 1               & 1                 & 1                  & 1               & 1                 & 1                  & 1               & 1                 & 1                  & 1               & 1                 & 43200                       \\ \hline
\textbf{Average}                                               & \BlackCell{1.00}               & \BlackCell{0.96}            & \BlackCell{0.98}              & \BlackCell{1.00}               & \BlackCell{0.97}            & \BlackCell{0.98}              & \BlackCell{1.00}               & \BlackCell{0.94}            & \BlackCell{0.96}              & \BlackCell{1.00}               & \BlackCell{0.95}            & \BlackCell{0.97}              & 345600                      \\ \hline
\end{tabular}
}
\caption{Anomaly Injection Perspective: Precision, Recall, and F-measure}
\label{tab:fmeasure-table-injection-master}
\end{table*}

\subsection{Efficacy} \label{sec:efficacy}

In this section we detail the efficacy of S-ESD and S-H-ESD using the
metrics mentioned earlier in this section. In particular, the following 
-- {\em precision, recall}, and {\em F-measure} -- described previously
in \ssecref{sec:fmeasure}, are reported.

\subsubsection{Perspectives} \label{sec:perspectives}

The performance of S-ESD and S-H-ESD was investigated from three different
perspectives -- 

(a) Capacity engineering (CapEng), 
(b) User behavior (UB) and 
(c) Supervised learning (Inj), wherein anomalies were injected in the input 
    time series to obtain labeled data. 

In the first two cases, service owners set a threshold that was then used to 
categorize the detected anomalies into true positives (TP) and false positives
(FP).

Specifically, 

\begin{dinglist}{114} 
\item The {\em CapEng} perspective was motivated by the fact that in capacity
      planning, a primary goal is to effectively scale the system resources 
      to handle the normal operating levels of the traffic, e.g., the expected 
      daily maximum, while maintaining enough headroom to absorb anomalies in
      input traffic.
      
      Thus, the objective in the current context is to detect anomalies that
      have a magnitude greater than a pre-specified threshold (which is in turn
      determined via load testing). 

\item From a user behavior (UB) perspective, the local intra-day anomalies serve
      as potential signals of change in user behavior. To this end, the service 
      owners set the threshold on the residual component of the time series of 
      interest.

      Note that setting a threshold as mentioned above is intrinsically directed 
      toward detection of only positive anomalies. In other words, anomalies in 
      the right tail of the underlying data distribution are detected. 

\item Lastly, the {\em Inj} perspective is aimed to assess the efficacy of the
      proposed techniques in the presence of ground-truth (or labeled data).
      Given the volume, velocity, and real-time nature of production data, it is
      not practically feasible to obtain labeled anomalies. To alleviate this limitation,
      we first fit a smooth spline (B-spline) curve to a time series (obtained from
      production) to derive a time series which had the same characteristics -- 
      trend and seasonality -- as of the original time series. Subsequently, 
      positive anomalies were injected in the derived time series, with varying
      magnitudes, widths, and frequency. The positions and magnitudes of the 
      injected anomalies were recorded and used to compare against the anomalies
      detected by both S-ESD and S-H-ESD.
\end{dinglist}

\subsubsection{Results} \label{sec:results}

We computed Precision, Recall, and F-measure (refer to \secref{sec:fmeasure})
to assess the efficacy of S-ESD and S-H-ESD from the three perspectives. 

Right-tailed (/positive) anomalies were detected at both 95\% and 99.9\% 
confidence levels. The metrics are reported in Tables \ref{tab:capeng} 
(CapEng), \ref{tab:ub} (UB), and \ref{tab:fmeasure-table-injection-master} 
(Inj). 

From the tables we note the following: across both system metrics and core 
drivers, precision increases from about 75\% in the CapEng perspective to 
100\% in the UB perspective for S-ESD, and about 59\% to 95\% for S-H-ESD. 
The threshold set in the CapEng perspective results in labeling of the 
intra-day or off-peak anomalies as false positives, thereby lowering the
precision. In contrast, in the UB perspective, the threshold is set for 
the residual component which removes the seasonality effect. This results
in S-H-ESD's higher recall rates, which improve from 47.5\% (S-ESD) to 77\% 
(S-H-ESD) for CapEng and from 31.5\% (S-ESD) to 65\% (S-H-ESD) for UB.  

Comparative analysis of the F-measure reported in Tables \ref{tab:capeng} and 
\ref{tab:ub} highlights that the F-measure is better in the latter case. 
This can be attributed, in part, to the fact that in case of the latter the
service-owners set the thresholds over the residual component (recall that
the time series decomposition removes the trend and seasonal component). 
This makes anomaly detection independent of the time at which they occurred,
e.g., on-peak on a daily max, or off-peak on a daily trough. The low values 
of F-measure in the CapEng perspective are reasoned at the end of this
section. 

\begin{figure}[!h]  
\centering 
\includegraphics[width=0.45\textwidth]{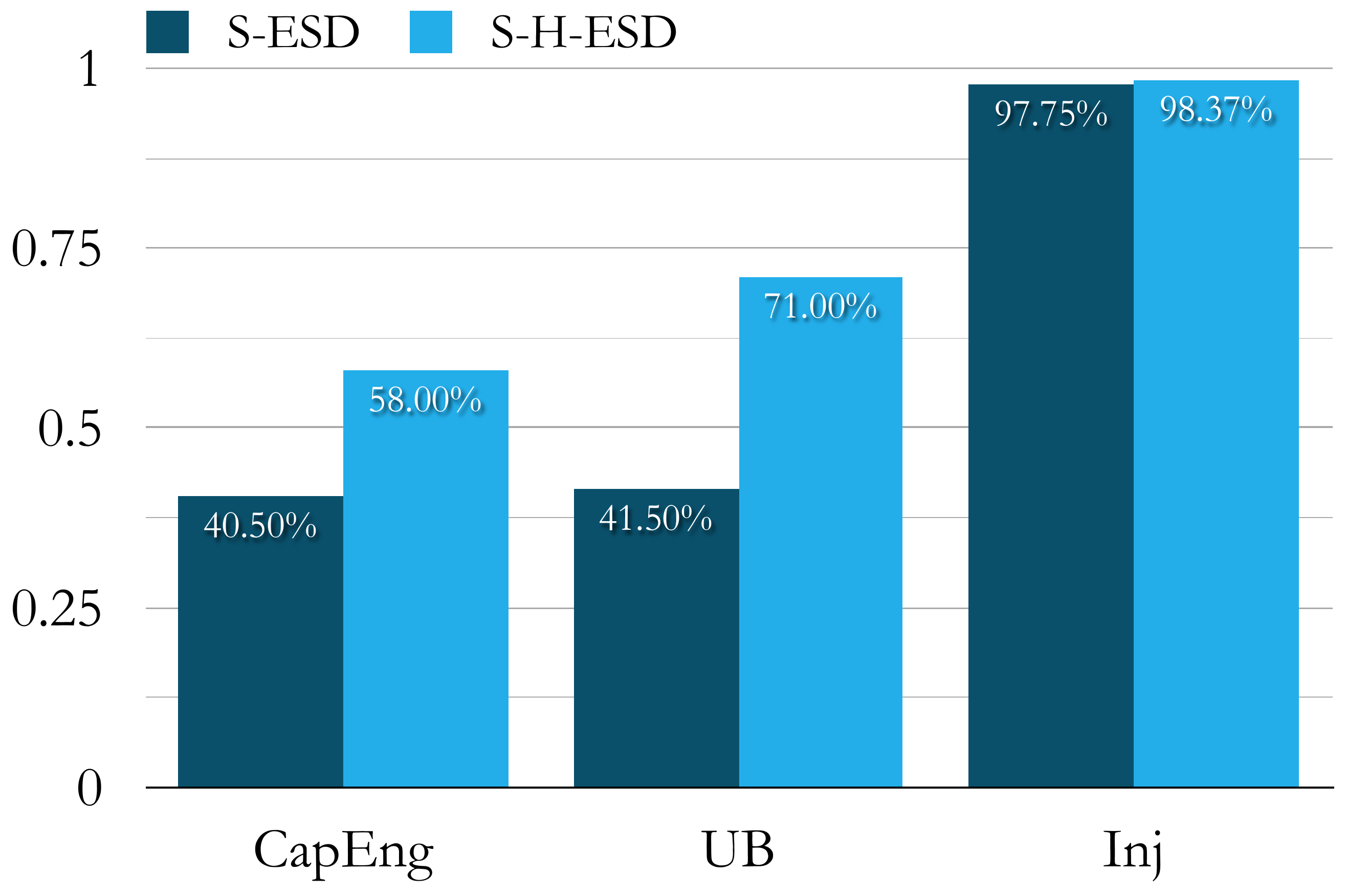}
\vspace*{-3mm}
\caption{F-measure: CapEng vs. UB vs. Inj (95\% Confidence)} 
\vspace*{-5mm}
\label{fig:fmeasurebarchart} 
\end{figure}

\noindent 
\tabref{tab:fmeasure-table-injection-master} reports the efficacy of S-ESD 
and S-H-ESD from the Inj (Ground-Truth) perspective. From the table we note
that the precision achieved was 100\%, meaning that all anomalies detected were 
{\em true} anomalies (there were no false positives). Also, the recall was 
very high, achieving about 96\% and 97\% (for S-ESD and S-H-ESD respectively)
at the 95\% confidence level and about 94\% and 95\% recall at the 99.9\% 
confidence level. 

On further analysis we noted that false negatives (anomalies that were not
detected) were within the boundaries of the anomalies detected, normally at 
the tail ends (at the beginning or end of the sustained anomalous behavior). 

\begin{figure}[!b]
\centering
\vspace*{-6mm}
\includegraphics[width=\linewidth]{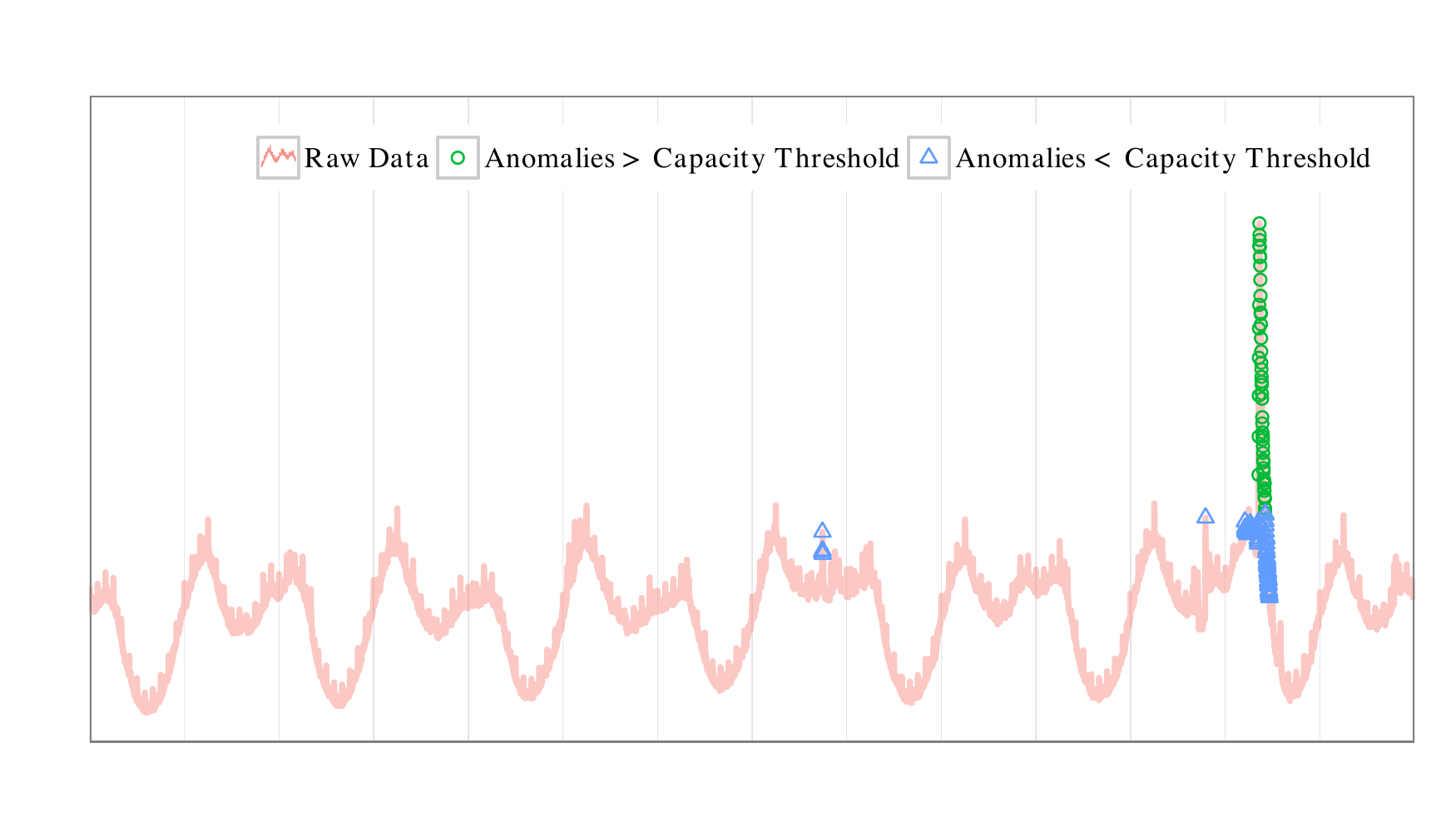}
\vspace*{-1cm}
\caption{Illustration of false positives in CapEng perspective}
\label{fig:capeng}
\end{figure}

Lastly, the results reported \tabref{tab:fmeasure-table-injection-master} 
correspond to injection sets containing anomalies with increasing magnitudes 
from $0.75\sigma$ to $6\sigma$. At $0.75\sigma$, the F-measure of S-ESD and 
S-H-ESD begins to degrade (0.84 (S-ESD) and 0.88 (S-H-ESD) at the 95\% confidence 
level and 0.76 (S-ESD) 0.82 (S-H-ESD) at the 99.9\% confidence level). At 
$1.5\sigma$, the F-measure was about 0.97 on an average. Injected anomalies 
with a magnitude of $3\sigma$ or greater achieved an F-measure of 1.00.

\figref{fig:fmeasurebarchart} summarizes the overall F-measure average for the
three perspectives at the 95\% confidence level. From the figure we note that 
in each case S-H-ESD outperformed S-ESD; in particular, the F-measure increased
by 17.5\%, 29.5\% and 0.62\% for CapEng, UB, and Inj respectively. This stems 
from the fact that median and MAD, unlike mean and standard deviation, are robust
against a large number of outliers. 

The F-measure for the CapEng and UB perspectives is significantly lower than 
the Inj perspective. This can be ascribed to the following: Capacity planning
engineers typically determine the capacity needed to withstand the typical
daily peaks (with additional headroom), thus the service-owners tend to set the
threshold around the maximum daily peaks. Consequently, anomalies which occur
during the off-peak hours of the day, such as in the daily troughs or other 
intra-day locations, would be marked as false positives (even though they might 
still be anomalous from S-ESD/S-H-ESD's point of view). This is illustrated in
\figref{fig:capeng} wherein the anomalies (detected using S-H-ESD) above the 
pre-specified threshold are annotated by \Circle \ and the rest are annotated 
by $\triangle$. The latter, albeit ``true'' anomalies from a statistical 
standpoint, are tagged as false positives which adversely impact precision
(refer to \eqnref{eq:prec}).

\section{Previous Work} \label{rel}

In this section, we overview prior work in the context of anomaly detection. 
A lot of anomaly detection research has been done in various domains such as, 
but not limited to, statistics, signal processing, finance, econometrics, 
manufacturing, and networking. For a detailed coverage of the same, the reader 
is referred to books and survey papers \cite{hawkins_identification_1980,
barnett_outliers_1994,hodge_survey_2004,aggarwal_outlier_2013}.

For better readability, we have partitioned this section into subsections on 
a per domain basis. As mentioned in a recent survey on anomaly detection
\cite{Chandola09}, anomaly detection is highly contextual in nature. Based
on our in-depth literature survey we find that the techniques discussed in 
the rest of this section cater to different type of data sets (than cloud 
infrastructure data) and hence are complementary to the techniques proposed
in this paper.  

\medskip
\noindent
{\normalsize \bf Manufacturing} 

\noindent
Anomaly detection manifests itself in manufacturing in the form of determining
if a particular process is in a state of normal and stable behavior. To this 
end, Statistical Process Control (SPC) was proposed in the early 1920s to monitor
and control the reliability of a manufacturing process \cite{shewhart_quality_1926,
Shewart31}. {\em Control charts} are one of the key tools used in SPC. 

In essence, the premise of SPC is that a certain amount of variation can occur
at any one point of a production chain. The variation is ``common" if it is
controlled and within normal or expected limits. However, the variation is
``assignable"  if it is not present in the causal system of the process at all
times (i.e., falls outside of the normal limits). Identifying and removing the
assignable sources which have impact on the manufacturing process is thus
crucial to ensure the expected operation and quality of the manufacturing
process.

A traditional control chart includes points representing a statistical
measurement (such as the mean) of a quality characteristic in samples taken over
a period of time. The mean of this characteristic is calculated over all
samples, and plotted as the center line. The standard deviation is calculated
over all samples, and the upper control limit (UCL) and lower control limit
(LCL) defined as the threshold at which the process is considered statistically
unlikely to occur (typically set at 3 standard deviations, denoted by $3
\sigma$, about the mean/center line). When the process is ``in control", 99.73\%
of all points are within the upper and lower control limits. A signal may
be generated when observations fall outside of these control limits, signifying the
introduction of some other source of variation outside of the normal expected
behavior.

Methodologies for improving the performance of control charts have been
investigated since Shewhart's early work. Roberts proposed the geometrical
moving average control chart -- also known as the EWMA (exponentially weighted
moving average) control chart -- which weights the most recent samples more
highly than older samples \cite{roberts_control_1959}. The EWMA chart tends to
detect small shifts (1-2 $\sigma$) in the sample mean more efficiently; however,
the Shewhart chart tends to detect larger shift (3 $\sigma$) more efficiently.
In case the quality characteristic follows a Poisson distribution, alternatives
to the EWMA chart have been proposed \cite{montgomery_introduction_2007}. Other 
types of control charts, such as the CUSUM (cumulative sum) chart
\cite{page_continuous_1954}, have been proposed. In \cite{lowry_review_1995},
Lowry and Montgomery present a review on control charts for multivariate quality
control. For further details on SPC and control charts, the reader is referred 
to the surveys by Woodall and Montgomery \cite{woodall_research_1999} and by 
Stoumbos et al.\ \cite{stoumbos_state_2000}. Lastly, a recent survey by Tsung et
al.\ presents a comprehensive survey of statistical control methods for multistage 
manufacturing and service operations \cite{tsung_statistical_2008}.

\medskip
\noindent
{\normalsize \bf Finance} 

\noindent
Behavioral economics and finance study the causal relationships between social, 
cognitive, and emotional factors on the economic decisions of individuals and 
institutions, and their effect on the economic decisions such as market prices 
and returns. 

The interplay of the aforementioned factors often result in, but not limited to,
seasonal stock market anomalies. These seasonal anomalies, also referred to as 
``calendar effects", take many forms. Haugen and Jorion describe the January
Effect -- stocks, especially small stocks, historically generate abnormally high
returns during the month of January -- as {\em ``... perhaps the best-known
example of anomalous behavior in security markets throughout the world"}
\cite{haugen_january_1996}. This effect is normally attributed to a theory which
states that the investors who hold a disproportionate amount of small stocks sell
their stocks for tax reasons at the end of the year (to claim capital loss), and
then reinvest at the start of the new year. On the other hand, the January effect 
is also attributed to the end of year bonuses which are paid in January and used 
to purchase stocks (thus driving up prices).

In \cite{angelovska_econometric_2013}, Angelovska reports that early studies on
stock market anomalies began in the late 1920's, where Kelly's reports
\cite{kelly_why_1930} showed the existence of a so-called {\em ``Monday effect"}
-- US markets have low and negative Monday returns. The Monday effect in the US
sock market was actively researched in the 1980s \cite{french_stock_1980}
\cite{gibbons_day_1981} \cite{rogalski_new_1984}. In
\cite{andrew_coutts_weekend_1999}, Coutts and Hayes showed that the Monday
effect exists albeit not as strongly as previous work demonstrated. Wang et al.\
show that between 1962 and 1993, the Monday effect is strongly evident in the
last two weeks of the month, for a wide array of stock indices. For a full
literature review on the Monday effect, refer to \cite{pettengill_survey_2003}.

Numerous studies support the existence of calendar effects in stock markets, as
well as others such as ``turn of the month" effects. For instance, Hensel and
Ziemba examined S\&P 500 returns over a 65 year period from 1928 to 1993, and
reported that U.S. large-cap stocks consistently show higher returns at the turn
of the month \cite{hensel_investment_1996}. In contrast, Sullivan et al.\ argue
against the case, stating that there is no statistically significant evidence
supporting the claim, and that such periodicities in stock market behavior are
the result of data dredging \cite{sullivan_dangers_2001}.  Subsequently, Hansen
et al.\ attempted to use sound statistical approaches to evaluate the
significance of calendar effects, by tightly controlling the testing to avoid
data mining biases \cite{hansen_testing_2005}. In their study of 27 stock
indices from 10 countries, calendar effects were found to be significant in most
returns series, with the end-of-the-year effect producing the largest anomalies and
the most convincing evidence supporting calendar effects in small-cap indices.

\medskip
\noindent
{\normalsize \bf Signal Processing} 

\noindent 
Techniques from signal processing such as, but not limited to, spectral analysis, 
have been adopted for anomaly detection. For instance, in \cite{cheng_use_2002},
Cheng et al.\ employed spectral analysis to complement existing DoS defense 
mechanisms that focus on identifying attack traffic, by ruling out normal TCP 
traffic, thereby reducing false positives. 

Similarly, wavelet packets and wavelet decomposition have been used for detecting
anomalies in network traffic \cite{barford_signal_2002,alarcon-aquino_anomaly_2001,
li_ddos_2005,kim_detecting_2004, ramanathan_wades:_2002}. Benefits of wavelet-based
techniques include the ability to accurately detect anomalies at various frequencies
(due to the inherent time-frequency property of decomposing signals into different
components at several frequencies), with relatively fast computation. 

In \cite{gao_anomaly_2006}, Gao et al.\ proposed a speed optimization for real-time
use using sliding windows. Recently, Lu et al.\ proposed an approach consisting 
of three components: (1) feature analysis, (2) normal daily traffic modeling based 
on wavelet approximation and ARX (AutoRegressive with eXogenous), and intrusion 
decision \cite{lu_network_2009}. An overview of signal processing techniques for 
network anomaly detection, including PSD (Power Spectral Density) and wavelet-based 
approaches, is presented in \cite{zhang_signal_2005}.

Additionally, Kalman filtering and Principle Component Analysis (PCA) based
approaches have been proposed in the signal processing domain for anomaly
detection. In \cite{ndong_signal_2011}, Ndong and Salamatian reported that
PCA-based approaches exhibit improved performance when coupled with
Karuhen-Loeve expansion (KL); on the other hand, Kalman filtering approaches,
when combined with statistical methods such as Gaussian mixture and Hidden
Markov models, outperformed the PCA-KL method.

\medskip
\noindent
{\normalsize \bf Network Traffic} 

\noindent
With the Internet of Things (IoT) paradigm \cite{Disruptive,Atzori10} increasingly 
becoming ubiquitous\footnote{According to a study by ABI Research \cite{IoT1}, it 
is estimated that 10 billion devices are currently connected to one another by
wired or wireless Internet. By the year 2020, that number is expected to exceed
30 billion.}, there is an increasing concern about security. In a post the FTC 
said on its website \cite{IoT}: ``At the same time, the data collection and 
sharing that smart devices and greater connectivity enable pose privacy and 
security risks.” Over the years, various anomaly detection techniques have been
proposed for detection network intrusion. 

For example, in \cite{denning_intrusion-detection_1987}, Denning proposed a 
rule-based technique wherein both network (system) and user data was used to 
detect different types of abnormal behavior, by comparing audit-trails to 
different anomalous profiles or models. 

In \cite{lazarevic_comparative_2003}, Lazarevic et al.\ presented a comparative
study of several anomaly detection schemes for network intrusion detection.

In \cite{garcia-teodoro_anomaly-based_2009}, Garcia-Teodoro et al.\ presented an
overview of the pros and cons of various approaches for anomaly detection in
network intrusion systems such as statistical techniques, knowledge-based
techniques (finite state machines, Bayesian networks, expert systems, etc.), and
learning based classification of patterns (Markov models, neural networks, fuzzy
logic, clustering, et cetera). Recently, Gogoi et al.\ presented a comprehensive
survey on outlier detection for network anomaly detection in \cite{gogoi_survey_2011}; 
in particular, the authors classified the approaches into three categories: 
(1) distance-based, (2) density-based, and (3) machine learning or soft-computing 
based. 

The reader is referred to the survey of intrusion detection techniques by Yu for
further reading \cite{Yu12}.

\medskip
\noindent
{\normalsize \bf Statistics} 

\noindent
Anomaly detection has been actively researched for over five decades in 
the domain of statistics \cite{Anscombe60,Bernoulli61,barnett_study_1978,
hawkins_identification_1980,Beckman83}. Recent surveys include the ones
from Hodge and Jim \cite{hodge_survey_2004} and Chandola et al.\ \cite{
chandola_anomaly_2009}. 

The key focus of prior work has been to determine whether a single value is
statistically anomalous with respect to an underlying distribution. Work by 
Markov and Chebyshev provided bounds on the probability of a random value
with respect to the expected value of the distribution. The Markov inequality
states that for any non-negative random variable X, the following holds true
$P(X > \alpha ) \leq E[X]/\alpha$, while the more general Chebyshev inequality
states that $P(|X - E[X]| > \alpha ) \leq Var[X]/\alpha^{2}$, and shows that
values equal to or greater then K standard deviations from the expected value
constitute no more then $1/k^{2}$ of the total distribution.

These bounds can be used as a threshold for determining the ``outlierness" of a
random value, indicating that a value does not fit the underlying distribution
\cite{barnett_outliers_1994}; however, the Markov and Chebyshev inequalities are
non-parametric, and create relatively weak bounds that may miss potential
outliers in the data \cite{aggarwal_outlier_2013}. The Chernoff bound and the
Hoeffding inequality attempt to create tighter bounds by making assumptions
about the underlying distribution. While these tail inequalities provided a 
closer bounds for testing the outlierness of a data point, their assumptions
regarding the distribution make them unsuitable for use when the distribution 
doesn't follow their underlying assumptions (as in the current context).  

Further, the {\em Box plot} may be applied as a robust means of determining if a 
data point is anomalous with respect to the underlying distribution \cite{
laurikkala_informal_2000}. A Box plot divides the data into five groups: the 
minimum non-anomalous value ({\em min}), the lower quartile ({\em Q1}), the 
median, the upper quartile ({\em Q3}) and the maximum non-anomalous value 
({\em max}). Data that is $1.5\times$ lower then ({\em Q1}) or $1.5\times$ 
greater then ({\em Q3}) are typically considered anomalous.

\section{Conclusion} \label{sec:conclusions} 

In this paper we presented two novel statistical techniques for automatically
detecting anomalies in cloud infrastructure data. Although there exists a large
body of research in anomaly detection, the seasonal (and trending) nature of 
cloud infrastructure data limits the application of techniques. To this end,
we proposed a method called \textit{Seasonal-ESD} (S-ESD), which combines
seasonal decomposition and the Generalized ESD test, for anomaly detection. 
The second method, \textit{Seasonal-Hybrid-ESD} (S-H-ESD), builds on S-ESD 
to enable robust anomaly detection when a significant portion (up to 50\%) of 
the underlying data is anomalous. This is achieved by extending the original 
ESD algorithm with robust statistical measures, median and median absolute 
deviation (MAD).

The efficacy of both S-ESD and S-H-ESD was evaluated using both core metrics
such as Tweets Per Sec (TPS), system metrics such as CPU and heap usage and 
application metrics. The evaluation was carried out from three different 
perspectives, viz., capacity engineering (CapEng), user behavior (UB), and 
supervised learning (Inj). Precision, Recall, and F-measure in each case.
Overall, S-H-ESD outperformed S-ESD, with F-Measure increasing by 17.5\%, 
29.5\% and 0.62\% for CapEng, UB, and Inj respectively. 

In light of the fact that S-H-ESD more computationally expensive than S-ESD 
(recall that the former requires sorting of the data), it is recommended to
use S-ESD in cases where the time series under consideration is large but 
with a relatively low anomaly count. 

As future work, we plan to extend the proposed techniques for detecting
anomalies in long time series. The challenge in this regard is that capturing
the underlying trend,\footnote{Capturing the trend is required to minimize the
number of false positives.} which in our observation is predominant in the case
of long time series, is non-trivial in the presenece of anomalies. To this end,
we plan to explore the use of qunatile regression \cite{Koenker78} and/or robust
regression \cite{Huber73,Rousseeuw03}.

{
\fontsize{5}{0.19cm}%
\selectfont%
\bibliographystyle{unsrt}%
\bibliography{bib/master}
}

\end{document}